\documentclass{styles/svproc}

\usepackage{url}
\usepackage{cite}
\usepackage{subcaption}
\usepackage{amsmath,amssymb,amsfonts}
\usepackage{algorithmic}
\usepackage{graphicx}
\usepackage{adjustbox}
\usepackage{textcomp}
\usepackage{xcolor}
\usepackage{makecell}
\usepackage{color}
\usepackage{graphicx}
\usepackage{float}    
\usepackage{wrapfig}
\usepackage{color}
\usepackage{array}
\usepackage{graphicx}
\usepackage{multirow}
\usepackage[table]{xcolor}
\usepackage{acronym}
\usepackage{graphicx}
\usepackage{subcaption}
\usepackage{booktabs}
\usepackage{multirow}
\usepackage{graphicx}
\usepackage{tabularx}
\usepackage{subcaption}
\usepackage{booktabs}
\usepackage{multirow}
\newacro{RMSE}{root mean square error}
\newacro{MAE} {mean absolute error}
\newacro{GP}{Gaussian process}
\newacro{GS}{Gaussian splatting}
\newacro{EDF}{Euclidean distance field}
\newacro{2DGS} {2D Gaussian splatting}
\newacro{3DGS} {3D Gaussian splatting}
\newacro{SDF} {signed distance field}
\newacro{DF} {distance field}
\newacro{UDF} {unsigned distance field}
\newacro{GBDF} {Gaussian basis distance field}
\newacro{GBE} {Gaussian basis element}
\newacro{CHOMP} {covariant Hamiltonian optimisation for motion planning}
\newacro{GPDF} {Gaussian process-based distance field}
\newacro{TSDF} {truncated signed distance field}
\newacro{GPIS} {Gaussian process implicit surface}
\newacro{GT} {ground-truth}

\begin{document}
\mainmatter              
\title{SplatlessDF: Continuous Distance Field Mapping with Non-Splatting Gaussians}
\titlerunning{SplatlessDF: Distance Field with Non-Splatting Gaussians}  

\author{Monisha Mushtary Uttsha\inst{1} \and Lan Wu\inst{2} \and
Teresa Vidal-Calleja\inst{1}}
\authorrunning{{M. Uttsha et al.}} 
\institute{
UTS Robotics Institute, Faculty of Engineering and IT,
University of Technology Sydney, Sydney, NSW 2007, Australia\\
\email{monishamushtary.uttsha@student.uts.edu.au}
\and
School of Engineering, University of Western Australia,
Perth, WA 6009, Australia
}

\maketitle              

\begin{abstract}

Recent \ac{GS} methods have shown that scenes can be represented efficiently with optimisable Gaussians for high-quality reconstruction and rendering. In this paper, building on this principle, we introduce \emph{SplatlessDF}, a continuous \ac{DF} mapping framework that uses anisotropic Gaussian elements from a spatial rather than photometric perspective. SplatlessDF directly parameterises the Gaussians and optimises to recover a differentiable \ac{DF}, enabling distances and gradients to be queried in the spatial domain for downstream robotic tasks such as navigation. Furthermore, SplatlessDF can be coupled with \ac{2DGS}, providing a unified framework based solely on Gaussian primitives that can learn continuous \ac{DF} and surface models and supports photometric rendering. We consider two settings: a standalone \ac{DF}-only formulation and a joint \ac{DF}--rendering formulation coupled with \ac{2DGS}. Experiments show that the standalone formulation provides efficient and accurate distance and gradient queries, while the joint formulation improves rendering geometry and simultaneously models a continuous \ac{DF}. These results highlight the potential of \ac{GS}-style representations not only for surface modelling and rendering but also for mapping representations suited to robotic navigation.

\end{abstract}

\section{Introduction}
\label{sec:introduction}
Robotic perception requires scene representations that support downstream geometric reasoning in addition to scene reconstruction and visual realism. In many robotic tasks, a useful map must provide quantities such as distance-to-surface, collision costs, and free-space structure in a form that can be queried efficiently for control, planning and navigation. Classical representations such as occupancy grids, \acp{SDF}, and Euclidean distance transforms therefore remain central to robotics, as they expose explicit spatial structure required for reliable geometric reasoning.

Recent advances in \acf{GS} have established Gaussian-based representations as an efficient and flexible paradigm for scene reconstruction and novel-view synthesis. In particular, \ac{3DGS} achieves high-fidelity rendering and real-time view synthesis by optimising anisotropic Gaussians directly in 3D space, avoiding the costly volumetric sampling used in neural radiance fields~\cite{3DGS-23}. Subsequent extensions have introduced stronger geometric structure through surfel-like formulations, depth and normal regularisation, and secondary \acf{DF} representations. For example, \acf{2DGS} constrains Gaussian primitives to oriented disks to improve geometric consistency~\cite{2DGS-24}, while recent hybrid methods such as GS-Octree~\cite{GSO-24}, SuGaR~\cite{SuGaR-24}, 3DGSR~\cite{3DGSR-24}, GS-SDF~\cite{GSSDF-25}, and PINGS~\cite{PINGS-25} combine Gaussian-based rendering with additional geometric representations to improve reconstruction, regularisation, or surface extraction.

However, a key gap remains between continuous \ac{DF} methods and \ac{GS}-based rendering methods. Existing approaches that recover continuous \acp{DF} typically rely on neural implicit~\cite{IGR-20,ISDF-22} or voxel-based~\cite{voxblox,vdbgpdf} formulations, where the field is learned or constructed independently of a \ac{GS}-inspired representation. Some of these methods, such as Voxblox~\cite{voxblox} and VDB-GPDF~\cite{vdbgpdf}, can provide geometric maps and coloured surface reconstructions, but they are not designed for photometric rendering or novel-view synthesis in the way \ac{GS} methods are. Conversely, recent hybrid methods such as PINGS~\cite{PINGS-25} and GS-SDF~\cite{GSSDF-25} demonstrate that \acp{DF} and Gaussian-based rendering can be optimised jointly, yet their \acp{DF} are not parameterised directly by Gaussians. As a result, the representational role of Gaussians remains primarily tied to rendering, rather than to the \ac{DF}.

In this paper, we address this gap by placing the \ac{DF} at the centre of the representation. We propose \emph{SplatlessDF}, a continuous \ac{DF} parameterised directly by anisotropic Gaussian primitives. Rather than using Gaussians as carriers of radiance or opacity, we treat them as continuous geometric elements whose aggregation defines a scalar \ac{DF} over Euclidean space. Although this follows the Gaussian-based modelling paradigm of \ac{GS}, the Gaussians are not splatted or rasterised on an image plane. Instead, because a \ac{DF} is evaluated directly at spatial coordinates, the formulation applies naturally to both 2D and 3D spatial domains. The model is trained using distance supervision derived from a point cloud and produces an accurate, continuously queryable \ac{DF} suitable for downstream robot navigation.

We develop this idea in two formulations. The first is a \emph{standalone formulation}, where SplatlessDF is optimised independently as a \ac{DF}-only map. This formulation targets the core robotic settings in which the required output is a continuous~\ac{DF} that can be queried for distances and gradients. The second is a \emph{joint formulation}, where the proposed \ac{DF} model is coupled with \ac{2DGS}~\cite{2DGS-24}, a Gaussian-based image renderer. In this case, \ac{2DGS} maintains its own separate set of Gaussian primitives, distinct in both parameters and purpose from those of the \ac{DF}. The two representations are therefore not merged into a single Gaussian set; instead, each is optimised with its own task-specific objective, while geometric-consistency losses couple the \ac{DF} and rendering representations. This joint formulation results in a framework with access to a continuous \ac{DF} representation for robotic applications, as well as improved rendering capabilities compared to the base \ac{2DGS} model. 

Overall, SplatlessDF provides a Gaussian-primitive formulation for accurate continuous \acp{DF} in both standalone and joint settings, supporting downstream navigation while additionally enabling image rendering in the joint formulation.

\section{Related Work}

\subsection{Continuous Distance Fields for Robotics}

Robotic \ac{DF} maps are commonly constructed from occupancy or \ac{TSDF} representations, with systems such as Voxblox~\cite{voxblox} demonstrating their utility for online reconstruction and planning. Learned implicit methods instead parameterise the \ac{DF} as a continuous function. DeepSDF~\cite{DeepSDF-19} learns \acp{SDF} from sampled signed-distance supervision, while IGR~\cite{IGR-20} shows that such fields can be learned directly from raw point clouds through geometric regularisation and surface constraints. More recent methods have broadened this landscape to both signed and unsigned settings, including iSDF~\cite{ISDF-22} for online scene reconstruction, CAP-UDF~\cite{CAP-UDF} for unsigned distance learning, and HotSpot~\cite{wang2025hotspot} for accurate signed distance optimisation.

\acp{GPIS} provide probabilistic continuous \ac{DF} representations. Bhoram et al.~\cite{Bhoram} perform online \ac{GPIS} mapping with local GP updates from noisy observations. Wu et al.~\cite{wu2023pseudo} optimise pseudo-input locations for accurate \acp{GPDF}, and VDB-GPDF~\cite{vdbgpdf} shows integration of \acp{GPDF} with OpenVDB~\cite{OpenVDB}, facilitating scalable online mapping. Unlike these neural and GP-based approaches, our method directly parameterises the \ac{DF} with optimisable Gaussian primitives, preserving continuous distance and gradient queries without neural implicit inference or GP kernel-matrix operations.

\subsection{Geometry-Aware Gaussian Splatting and Distance Fields}

\ac{GS} represents radiance fields with anisotropic Gaussian primitives and enables efficient differentiable rendering~\cite{3DGS-23}. Geometry-aware variants improve the surface structure of Gaussian representations: \ac{2DGS} uses oriented planar disks for better surface alignment, depth, and normal estimation~\cite{2DGS-24}, while surface-aligned methods improve mesh extraction and geometric consistency~\cite{SuGaR-24}. These methods improve rendering-oriented geometry, but do not directly provide a continuous \ac{DF} for distance queries.

GS-Octree~\cite{GSO-24} uses an octree-based implicit surface to guide Gaussian optimisation. 3DGSR~\cite{3DGSR-24} and GaussianRoom~\cite{GR-25} jointly optimise neural \acp{SDF} with Gaussian primitives, while GS-SDF~\cite{GSSDF-25} adds LiDAR-informed neural \ac{SDF} constraints for consistent rendering and reconstruction. PINGS~\cite{PINGS-25} couples a \ac{DF} and a \ac{GS} radiance field within a point-based implicit neural map for LiDAR-visual SLAM. The methods typically represent \ac{DF} by a neural, voxel-based, or point-based secondary model, while reserving Gaussian primitives primarily for rendering; in contrast, we directly use Gaussian primitives as the \ac{DF} parameterisation itself and further couple this field with Gaussian-based rendering.

\section{Problem Setup and Background}
\label{sec:prelim_problem}

\subsection{Problem Setup}
\label{subsec:problem_setup}

We seek a continuous geometric representation of a scene that supports continuous distance queries suitable for downstream robotic tasks such as planning and navigation. Let $\mathcal{P}=\{p_j\}_{j=1}^{N}\subset\mathbb{R}^{d}$ denote a registered point cloud of surface observations, with $d\in\{2,3\}$. We assume that $\mathcal{P}$ provides sufficient coverage of the observed scene geometry, and aim to learn a continuous \ac{EDF} $D_\theta:\mathbb{R}^{d}\rightarrow\mathbb{R}_{\geq 0}$ that maps any query point $x\in\mathbb{R}^{d}$ to its distance from the nearest surface.

When image rendering capabilities are required, we additionally consider RGB observations $\mathcal{I}=\{(I_k,\Pi_k)\}_{k=1}^{K}$, where $I_k$ denotes an image and $\Pi_k$ its associated camera pose. In this setting, the objective is to learn a continuous \ac{EDF} while jointly optimising an image-rendering representation of the same scene to attain additional rendering capabilities.
\subsection{Background: 2D Gaussian Splatting}
\label{sec:prelim_2dgs}

We briefly review \ac{2DGS}~\cite{2DGS-24}, which is used later as the rendering backbone in our joint formulation. \ac{2DGS} represents a scene as a set of oriented planar Gaussian disks, or equivalently, 2D Gaussian surface primitives, embedded in 3D space.
Let $\psi$ denote the learnable parameters of this representation. 
For the $i$-th disk, these parameters include its centre $c_i(\psi)\in\mathbb{R}^{3}$, two local tangent directions $t_{1,i}(\psi),t_{2,i}(\psi)\in\mathbb{R}^{3}$, in-plane scales $s_{1,i}(\psi),s_{2,i}(\psi)$, opacity $\alpha_i(\psi)$, and view-dependent appearance coefficients. 
 
A local coordinate $(u,v)$ on the disk is mapped to 3D by
\begin{equation}
p_i(u,v;\psi)
=
c_i(\psi)
+
u\,s_{1,i}(\psi)t_{1,i}(\psi)
+
v\,s_{2,i}(\psi)t_{2,i}(\psi).
\label{eq:2dgs_disk_param}
\end{equation}
The contribution of each disk is weighted by a Gaussian kernel in this local tangent plane. Rendering is performed by intersecting each camera ray with the Gaussian disks, evaluating their local Gaussian weights, and compositing
their colour contributions using front-to-back alpha blending.
Compared with volumetric 3DGS's~\cite{3DGS-23} Gaussians, the planar primitives of \ac{2DGS} are more closely aligned with surfaces and provide an explicit local surface parameterisation. 
Given RGB images $\mathcal{I}$, the standard \ac{2DGS} objective combines photometric reconstruction with its geometry-aware regularisation terms:
\begin{equation}
\mathcal{L}_{\mathrm{2DGS}}(\psi)
=
\mathcal{L}_{\mathrm{photo}}(\psi)
+
\lambda_{\mathrm{depth}}\mathcal{L}_{\mathrm{depth}}(\psi)
+
\lambda_{\mathrm{norm}}\mathcal{L}_{\mathrm{norm}}(\psi),
\label{eq:2dgs}
\end{equation}
where $\mathcal{L}_{\mathrm{photo}}$ is the RGB reconstruction loss, while
$\mathcal{L}_{\mathrm{depth}}$ and $\mathcal{L}_{\mathrm{norm}}$ denote the depth-distortion and normal-consistency regularisers. 
We retain this rendering formulation unchanged and introduce additional coupling losses with the proposed Gaussian-based \ac{DF} in Sec.~\ref{sec:joint_optimisation}.

\section{Methodology}
\label{sec:method}

\begin{figure}[t]
    \centering
    \begin{subfigure}[t]{\columnwidth}
        \centering
        \fbox{%
            \includegraphics[width=0.95\columnwidth,height=0.23\columnwidth]{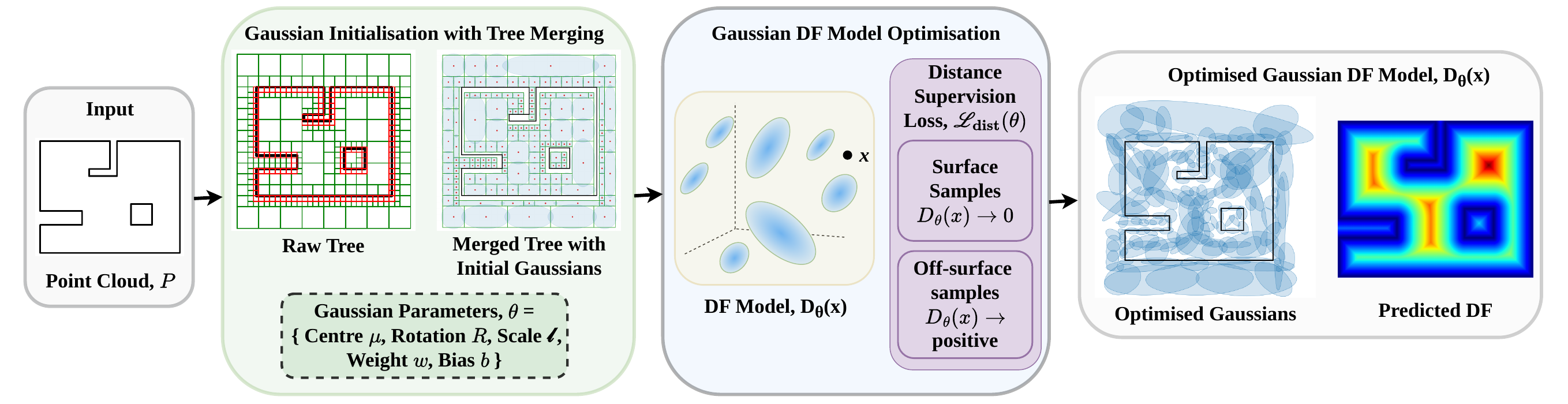}%
        }
        \caption{Standalone Gaussian \ac{DF} modelling pipeline.}
        \label{fig:gbdf_modelling_pipeline}
    \end{subfigure}

    \vspace{0.2em}

    \begin{subfigure}[t]{\columnwidth}
        \centering
        \fbox{%
            \includegraphics[width=0.95\columnwidth,height=0.28\columnwidth]{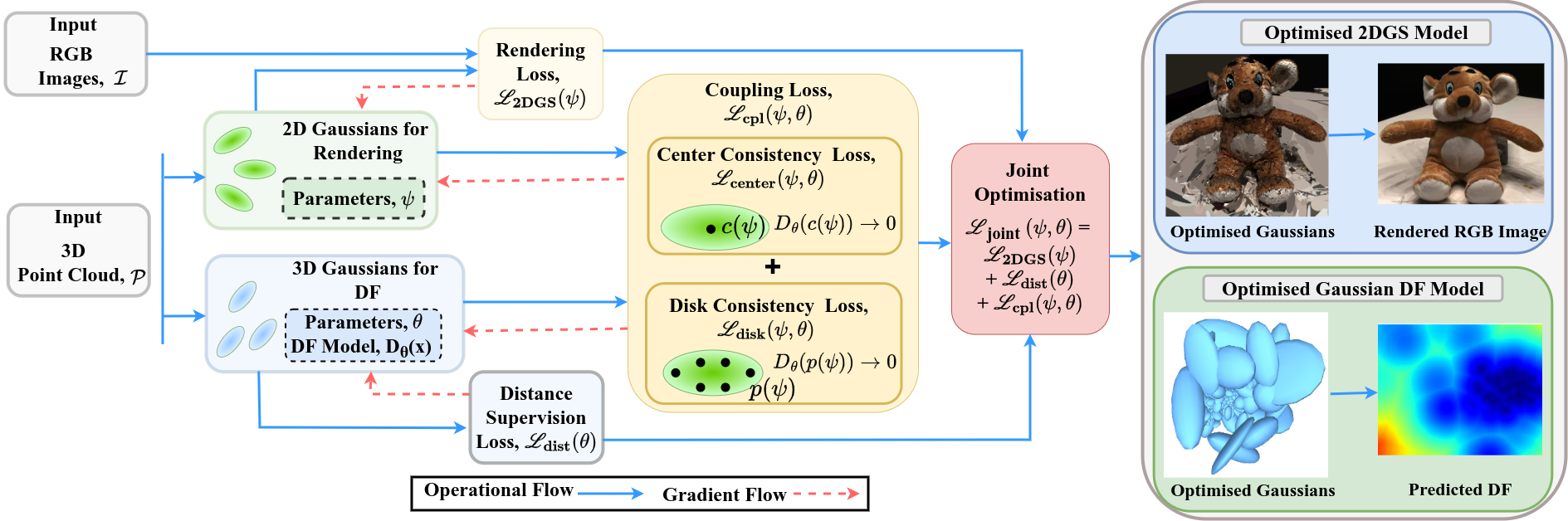}%
        }
        \caption{Joint optimisation pipeline of Gaussian \ac{DF} and \ac{2DGS}.}
        \label{fig:gbdf_joint_pipeline}
    \end{subfigure}

    \caption{
    Overview of the proposed pipelines.
    }
    \label{fig:gbdf_pipelines}
\end{figure}

\subsection{Overview}
\label{subsec:method_overview}

This section describes the proposed SplatlessDF and its two formulations, illustrated in Figure~\ref{fig:gbdf_pipelines}. In the standalone formulation of Figure~\ref{fig:gbdf_modelling_pipeline}, the input point cloud $\mathcal{P}$ is used to build a multi-resolution tree over the scene. Neighbouring cells with similar nearest-surface distances are merged, and each retained region initialises a Gaussian. The resulting Gaussian primitives parameterise a continuous \ac{DF} $D_\theta(x)$, which is optimised from $\mathcal{P}$ using surface samples with $D_\theta(x)\rightarrow 0$ and off-surface samples with positive nearest-surface distances. This yields an optimised \ac{DF} that supports continuous distance and gradient queries.

In the joint formulation of Figure~\ref{fig:gbdf_joint_pipeline}, given RGB images $\mathcal{I}$, train a \ac{2DGS} renderer with parameters $\psi$, while $\mathcal{P}$ trains a separate set of Gaussians for \ac{DF}, with parameters $\theta$. 
The two representations use distinct Gaussian primitives, and each is supervised by its own task-specific objective: $\mathcal{L}_{\mathrm{2DGS}}(\psi)$ for rendering and $\mathcal{L}_{\mathrm{dist}}(\theta)$ for \ac{DF} learning. In addition, differentiable centre- and disk-consistency losses jointly update the two representations, encouraging the rendering geometry to agree with the zero-distance structure of the \ac{DF}. The resulting joint objective therefore produces two task-specific but geometrically coupled outputs: an optimised \ac{2DGS} model for rendering and an optimised \ac{DF} model for distance-related queries.
 
\subsection{Euclidean Distance Field with Non-splatting Gaussians}
\label{sec:gbdf}

We model the \ac{DF} as a continuous function parameterised by trainable anisotropic Gaussians in Euclidean space. Unlike rendering-oriented \ac{GS}, these Gaussians are not projected, rasterised, or alpha-composited; the field is evaluated directly at each spatial query point. The formulation applies in both 2D and 3D, and is written below for $\mathbb{R}^{d}$ with $d\in\{2,3\}$.

Let the model contain $M$ Gaussian primitives. The $i$-th element is parameterised by a mean $\mu_i \in \mathbb{R}^{d}$, a rotation $R_i \in SO(d)$, a positive local scale vector $\ell_i \in \mathbb{R}^{d}_{+}$, and a scalar weight $w_i \in \mathbb{R}$. 
Its covariance is given by $\Sigma_i = R_i \, \mathrm{diag}(\ell_i^2)\, R_i^\top$.
For a query point $x\in\mathbb{R}^{d}$, the predicted unsigned distance is defined as
\begin{equation}
D_\theta(x)
=
\mathrm{softplus}\!\left(
\sum_{i=1}^{M} w_i \phi_i(x) + b
\right),
\label{eq:gpdf_field_merged}
\end{equation}
where
\begin{equation}
\phi_i(x)
=
\exp\!\left(
-\frac{1}{2}(x-\mu_i)^\top \Sigma_i^{-1}(x-\mu_i)
\right),
\end{equation}
$b\in\mathbb{R}$ is a learnable bias, and
$\theta=\{\mu_i,R_i,\ell_i,w_i\}_{i=1}^{M}\cup\{b\}$ denotes the trainable parameters. The softplus activation yields positive Euclidean distance predictions that approach zero near the observed surface. Supervision is obtained from discrete distance samples approximated from the observed scene geometry. Let $\Omega \subset \mathbb{R}^{d}$ denote the bounded training domain, defined as the bounding box of the observed point cloud $\mathcal{P}$ with a small padding margin. We then sample discrete training locations $\mathcal{X}=\{x_m\}_{m=1}^{N_X}$, where $x_m \in \Omega$, and assign each location its unsigned Euclidean distance to the observed point cloud $\tilde d_m = \min_{q\in\mathcal{P}}\|x_m-q\|_2$.
This gives the estimated supervision set $\tilde{\mathcal{X}} = \{(x_m,\tilde d_m)\}_{m=1}^{N_X}$.

The target distances $\tilde d_m$ are efficiently computed once using KD-tree nearest-neighbour queries on $\mathcal{P}$ and are then reused during optimisation. 
At iteration $t$, a minibatch $\mathcal{B}_t \subset \tilde{\mathcal{X}}$ is sampled and the model is optimised using
\begin{equation}
\mathcal{L}_{\mathrm{dist}}(\theta)
=
\frac{1}{|\mathcal{B}_t|}
\sum_{(x,\tilde d)\in\mathcal{B}_t}
\mathrm{smooth}_{\ell_1}\!\left(D_\theta(x)-\tilde d\right),
\label{eq:dist_loss_merged}
\end{equation}
where $\mathrm{smooth}_{\ell_1}$ is the standard Smooth L1 penalty.
Since Eq.~\eqref{eq:gpdf_field_merged} defines a differentiable function, spatial gradients are available directly from the learned field:
\begin{equation}
\nabla_x D_\theta(x)
=
\sigma\!\left(
\sum_{i=1}^{M} w_i\phi_i(x)+b
\right)
\sum_{i=1}^{M} w_i \nabla_x \phi_i(x),
\end{equation}
with
\begin{equation}
\nabla_x \phi_i(x)
=
-\phi_i(x)\Sigma_i^{-1}(x-\mu_i),
\end{equation}
where $\sigma(\cdot)$ denotes the logistic sigmoid. 
Thus, although supervision is imposed only at sampled training locations, the learned model remains a continuous \ac{DF} that can be queried at arbitrary points in $\Omega$.
This property is important for robotics, where collision checking, trajectory optimisation, and navigation-cost computation require repeated distance and gradient queries.

\subsection{Joint Optimisation with 2DGS}
\label{sec:joint_optimisation}
 
We now describe the geometric coupling used in the joint formulation. 
The \ac{DF} model and the \ac{2DGS} renderer interact through losses applied to reliable \ac{2DGS} disks. Since these disks represent local surface elements, their centres and sampled disk points should lie near low-distance regions of the learned \ac{DF}.

For these losses, we select a reliable subset of \ac{2DGS} disks. Let $\mathcal{G}=\{1,\ldots, N_{\psi}\}$ denote the indices of all \ac{2DGS} disks, and let $\mathcal{C}\subset\mathcal{G}$ denote the subset selected for coupling at the current iteration. A disk is considered reliable if it is visible in the current training view and its opacity satisfies $\alpha_i(\psi) \geq \alpha_{\min}$, where $\alpha_{\min}$ is an opacity threshold. For efficiency, we cap the subset size such that $|\mathcal{C}|\leq K_c$ per iteration. The same subset is used for both losses.

\subsubsection{Centre consistency loss.}
Let $c_i(\psi)\in\mathbb{R}^3$ denote the centre of a selected \ac{2DGS} disk, with $i\in\mathcal{C}$. We evaluate the Gaussian-based \ac{DF} directly at these centres and penalise their predicted distances:

\begin{equation}
\mathcal{L}_{\mathrm{center}}(\psi,\theta)
=
\frac{1}{|\mathcal{C}|}
\sum_{i\in\mathcal{C}}
D_{\theta}\!\left(c_i(\psi)\right).
\label{eq:lcenter}
\end{equation}
This term encourages the \ac{DF} to assign low distance values to reliable \ac{2DGS} disk centres, while the gradient through $c_i(\psi)$ also encourages those centres to move toward low-distance regions of the learned field.
\subsubsection{Disk consistency loss.}
Centre consistency only constrains the disk centre and does not directly regularise the local disk shape. Following the idea of shape regularisation in \cite{GSSDF-25}, we additionally sample points on the plane of selected \ac{2DGS} disks and encourage these points to lie close to the zero-distance set of the learned \ac{DF}. For the $i$-th selected disk, let $\mathcal{S}_i(\psi)$ denote the set of sampled points on the local disk, and let $p\in\mathcal{S}_i(\psi)$ denote one such sampled point. The disk consistency loss is defined as

\begin{equation}
\mathcal{L}_{\mathrm{disk}}(\psi,\theta)
=
\frac{1}{N_s}
\sum_{i\in\mathcal{C}}
\sum_{p\in\mathcal{S}_i(\psi)}
D_{\theta}\!\left(p\right),
\label{eq:lsupport}
\end{equation}
where $N_s$ is the number of sampled points per disk. Since $D_\theta(\cdot)$ is non-negative, minimising Eq.~\eqref{eq:lsupport} encourages the sampled disk points to lie close to the observed surface. Gradients from this term propagate through the sampled points, and according to Eq.~\eqref{eq:2dgs_disk_param}, influence the disk centre, scale, and tangent frame.
The full coupling loss is
\begin{equation}
\mathcal{L}_{\mathrm{cpl}}(\psi,\theta)
=
\lambda_{\mathrm{center}}\mathcal{L}_{\mathrm{center}}(\psi,\theta)
+
\lambda_{\mathrm{disk}}\mathcal{L}_{\mathrm{disk}}(\psi,\theta).
\label{eq:lcpl}
\end{equation}
The overall joint objective is
\begin{equation}
\mathcal{L}_{\mathrm{joint}}(\psi,\theta)
=
\mathcal{L}_{\mathrm{2DGS}}(\psi)
+
\mathcal{L}_{\mathrm{dist}}(\theta)
+
\mathcal{L}_{\mathrm{cpl}}(\psi,\theta),
\label{eq:ljoint}
\end{equation}
where $\mathcal{L}_{\mathrm{2DGS}}$ is the rendering objective from Eq.~\eqref{eq:2dgs} and $\mathcal{L}_{\mathrm{dist}}$ is the distance-supervision loss in Eq.~\eqref{eq:dist_loss_merged}. 
Thus, each representation keeps its own task-specific objective, while the coupling terms provide interaction between the \ac{2DGS} primitives and the Gaussian-based \ac{DF} model.

\subsection{Gaussian Centre Initialisation for Distance Field Modelling}
\label{subsec:meth_gauss_init}
 For both the formulations, we initialise the set of \ac{DF}-Gaussians in a geometry-aware manner rather than at random. Given the reference point cloud, \(\mathcal{P}\), we first construct a quadtree (\(d=2\)) or octree (\(d=3\)) over the scene domain and collect the centres of empty leaf cells. For each empty leaf cell centre \(c_\ell\), we compute its unsigned distance to surface $d_\ell=\min_{q\in\mathcal{P}}\|c_\ell-q\|_2$ using KD-tree based nearest-neighbour distance queries on \(\mathcal{P}\). Candidate cells that are adjacent in the tree structure and whose centres' distance values differ by less than a merging threshold $\tau_m$ are merged to form coarser regions in geometrically uniform areas. The final set of retained regions therefore consists of both merged regions and empty leaf cells that were not merged. If \(\widetilde{r}_j\) denotes one such retained region, with centroid \(\tilde{c}_j\) and size \(\tilde{s}_j\), we initialise a Gaussian by setting its mean to \(\mu_j=\tilde{c}_j\) and its scale from \(\tilde{s}_j\). This yields a scene-adaptive multi-resolution Gaussian set, with smaller Gaussians in higher distance variation regions and larger ones in homogeneous free-space. This tree-based procedure is used only to provide an informed initialisation of the primitive parameters, which are subsequently tuned through optimisation.
\section{Experimental Results}

\subsection{Experimental Setup}

\paragraph{\textbf{Datasets:}}

We evaluate the proposed method in both 2D and 3D settings. For 2D \ac{DF} evaluation, we use the synthetic Snowflake shape ($1.2\,\mathrm{m} \times 1.2\,\mathrm{m}$) from~\cite{wang2025hotspot} and the larger Gazebo~\cite{Bhoram} dataset ($20\,\mathrm{m} \times 16\,\mathrm{m}$), which contains LiDAR scans of a simulated environment. For 3D evaluation, we use two RGB-D indoor sequences: \emph{Freiburg 3 long office household} from the TUM dataset and Cow\&Lady. The TUM sequence provides synchronised RGB-D data and \ac{GT} trajectories at $30\,\mathrm{Hz}$ with $640 \times 480$ image resolution. Cow\&Lady provides Kinect RGB-D images at $640 \times 480$ resolution with Vicon-tracked poses, and is challenging due to fast camera motion, large viewpoint changes, and pose misalignment. For Cow\&Lady, to avoid biasing the rendering evaluation toward the initial viewpoint, we discard the first 70 near-duplicate frames.

\paragraph{\textbf{Benchmarks and Evaluation Setup:}}
Our evaluation is organised around four aspects of performance: continuous \ac{DF} prediction, rendering quality, surface reconstruction, and downstream navigation utility. For \ac{DF} evaluation, we benchmark the proposed Gaussian \ac{DF} against HotSpot~\cite{wang2025hotspot}, CAP-UDF~\cite{CAP-UDF}, and \ac{GPDF} from~\cite{wu2023pseudo}. We evaluate all methods in the unsigned setting, comparing Euclidean distance-to-surface values. Since HotSpot generates signed distance, we take the absolute value. \ac{DF} accuracy is reported using \ac{RMSE}; gradient quality is evaluated using mean cosine similarity for direction and \ac{MAE} of the gradient-norm deviation from the Euclidean-distance condition $\|\nabla D\|=1$ for magnitude. All distance-related quantities are reported in metres (m), unless otherwise specified. These comparisons include both our standalone (SA) and joint (Joint) formulations. 

The proposed joint formulation's rendering quality is assessed by comparing with vanilla \ac{2DGS} against \ac{GT} RGB images, using PSNR, SSIM, LPIPS, and L1 error. For surface reconstruction, we evaluate meshes extracted from all the methods, including vanilla \ac{2DGS}, reporting Chamfer-L1 distance. To validate the utility of the proposed Gaussian \acp{DF} for robotics, we further evaluate navigation on the 3D scenes using \ac{CHOMP}, which requires reliable distance and gradient queries for collision-free trajectory optimisation. We compare against the best-performing benchmark among the evaluated 3D \ac{DF} methods on safety-aware navigation. All experiments are run on an Intel Xeon Gold 6126 CPU and an NVIDIA Tesla V100 GPU with 32~GB memory.

\subsection{Scene-Aware Gaussian Budget Selection}
\label{sec:res_gaussian_budget_selection}

The Gaussian budget of our \ac{DF} representations follows from the multi-resolution tree construction and merging strategy in Section~\ref{subsec:meth_gauss_init}. 
By merging neighbouring cells, the method controls the number of retained Gaussians in a geometry-aware manner, rather than fixing the primitive count manually.

Table~\ref{tab:gaussian_budget_init_ablation} shows that the proposed merged-tree initialisation reduces the number of Gaussians compared with the non-merged raw tree (green shaded columns) while maintaining comparable distance accuracy. When compared with uniform-grid initialisation (grey shaded columns) at matched budgets, it also gives stronger gradient behaviour without degrading distance accuracy, indicating that the gain comes from geometrically meaningful placement of the Gaussians rather than primitive count alone. Across the tested settings, increasing the Gaussian budget generally improves distance accuracy and gradient agreement, although the gains become modest beyond an intermediate budget. The underlined budgets correspond to distance-merging thresholds of $\tau_m=0.05$\,cm, $0.5$\,cm, $1$\,cm, and $1$\,cm for Snowflake, Gazebo, TUM, and Cow\&Lady, respectively, and are used as defaults in the remaining experiments.
 
\begin{table*}[t]
\centering
\caption{Effect of Gaussian budget and initialisation strategy on proposed \ac{DF} mapping. 
Green-shaded columns report the raw tree initialisation, unshaded columns report the proposed merged-tree initialisation, and underlined budgets denote the selected budgets used in the main experiments. Columns marked $\dagger$ and shaded grey report uniform-grid initialisation at the same selected budget.}
\label{tab:gaussian_budget_init_ablation}
\resizebox{1\textwidth}{!}{%
\begin{tabular}{
l
>{\columncolor{green!12}}c c >{\columncolor{gray!12}}c c c|
>{\columncolor{green!12}}c c >{\columncolor{gray!12}}c c c|
>{\columncolor{green!12}}c c >{\columncolor{gray!12}}c c c|
>{\columncolor{green!12}}c c >{\columncolor{gray!12}}c c c
}
\hline
\multirow{2}{*}{Metric} 
& \multicolumn{5}{c|}{Snowflake} 
& \multicolumn{5}{c|}{Gazebo} 
& \multicolumn{5}{c|}{TUM (SA)} 
& \multicolumn{5}{c}{Cow\&Lady (SA)}\\ 
\cline{2-21}
& 200 & \underline{100} & $100^\dagger$ & 50 & 15
& 600 & \underline{400} & $400^\dagger$ & 200 & 40
& 1750 & \underline{1350} & $1350^\dagger$ & 1050 & 800
& 3200 & \underline{1600} & $1600^\dagger$ & 1000 & 700 \\
\hline
\ac{RMSE} $\downarrow$
& 0.004 & 0.005 & 0.005 & 0.008 & 0.015
& 0.016 & 0.019 & 0.017 & 0.030 & 0.090
& 0.017 & 0.020 & 0.024 & 0.024 & 0.026
& 0.030 & 0.035 & 0.036 & 0.046 & 0.052 \\
Mean cos. sim. $\uparrow$
& 0.96 & 0.94 & 0.93 & 0.86 & 0.78
& 0.98 & 0.97 & 0.92 & 0.96 & 0.96
& 0.92 & 0.90 & 0.85 & 0.87 & 0.82
& 0.88 & 0.84 & 0.82 & 0.82 & 0.80 \\
\ac{MAE}$(\|\nabla D\|-1)$ $\downarrow$
& 0.09 & 0.09 & 0.14 & 0.19 & 0.29
& 0.06 & 0.09 & 0.13 & 0.11 & 0.17
& 0.04 & 0.04 & 0.09 & 0.05 & 0.05
& 0.12 & 0.14 & 0.14 & 0.17 & 0.18 \\
\hline
\end{tabular}%
}
\end{table*}

\subsection{Distance Field Accuracy and Gradient Quality}

For our \acp{DF}, supervision is estimated at mini-batched query points randomly sampled from a uniform spatial grid. The 2D standalone model uses grids with $0.1\,\mathrm{m}$ spacing, 100 query points per iteration, and $15$k iterations. In 3D, both formulations use 3D grids with $0.2\,\mathrm{m}$ spacing, 500 query points per iteration, and $30$k iterations.

\begin{figure*}[t]
    \centering
    \setlength{\tabcolsep}{1pt}

    \begin{tabular}{ccccc}
        
        \includegraphics[width=0.18\textwidth, height=0.15\columnwidth]{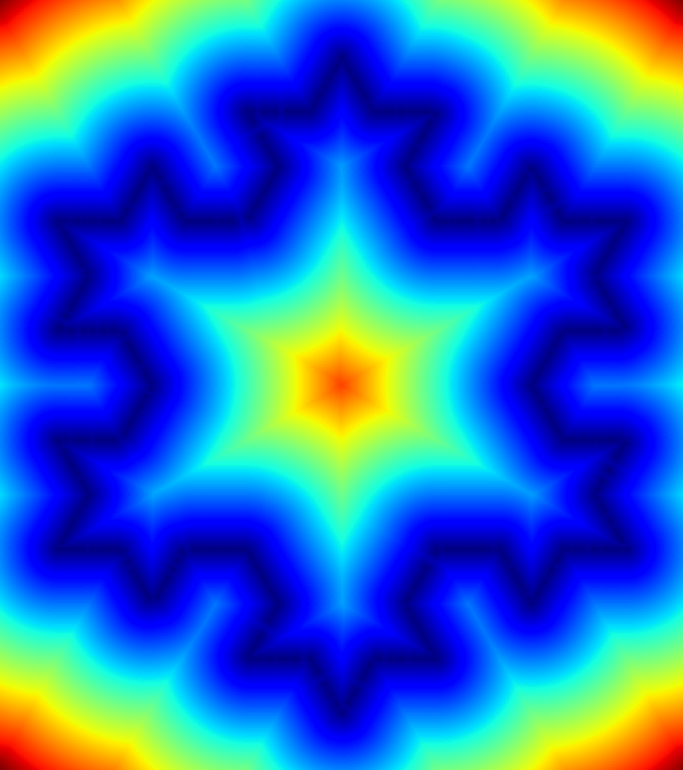} &
        
        \includegraphics[width=0.18\textwidth, height=0.15\columnwidth]{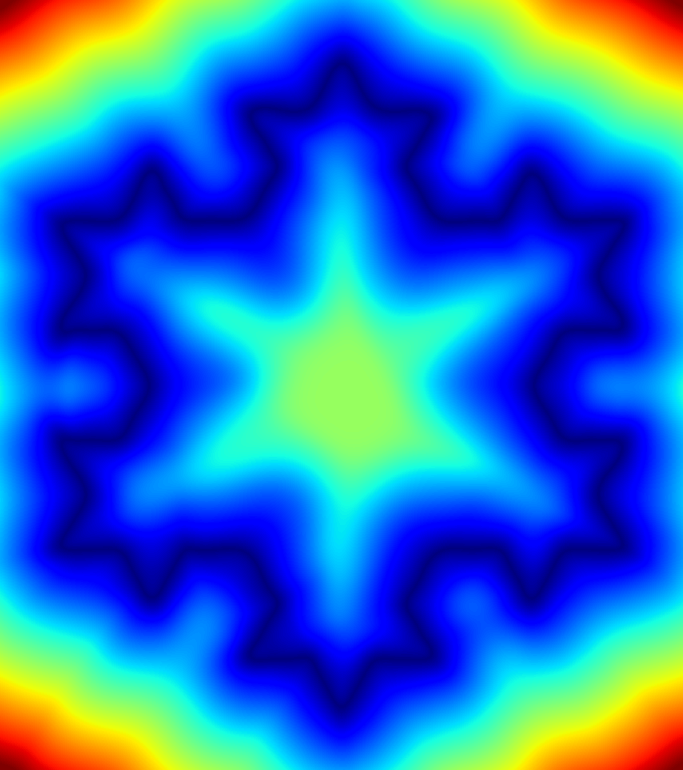} &
        \includegraphics[width=0.18\textwidth, height=0.15\columnwidth]{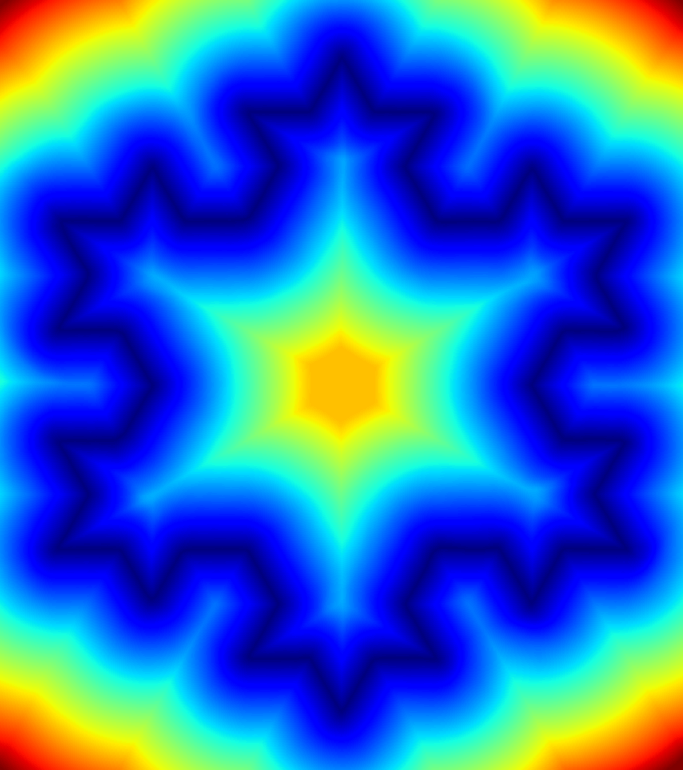} &
        \includegraphics[width=0.18\textwidth, height=0.15\columnwidth]{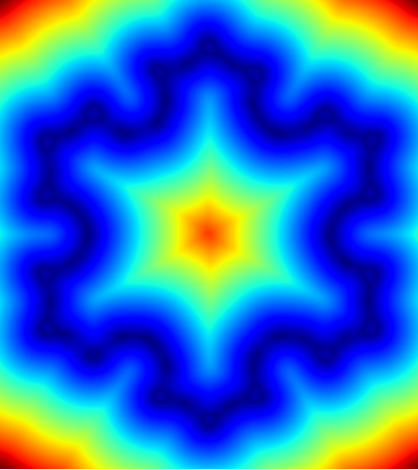} &
        \includegraphics[width=0.218\textwidth, height=0.15\columnwidth]{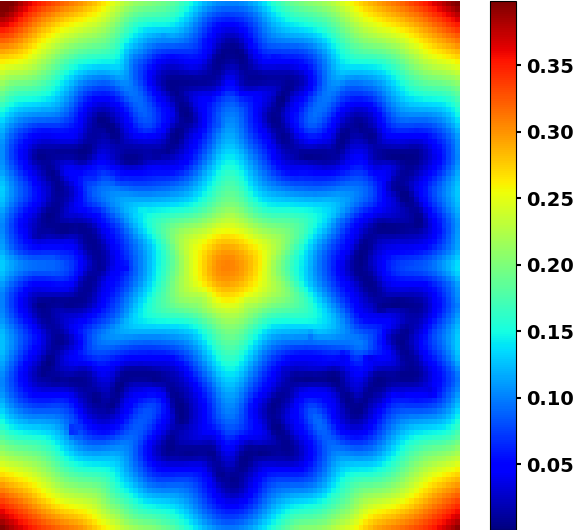} \\[0.2em]

        \includegraphics[width=0.18\textwidth]{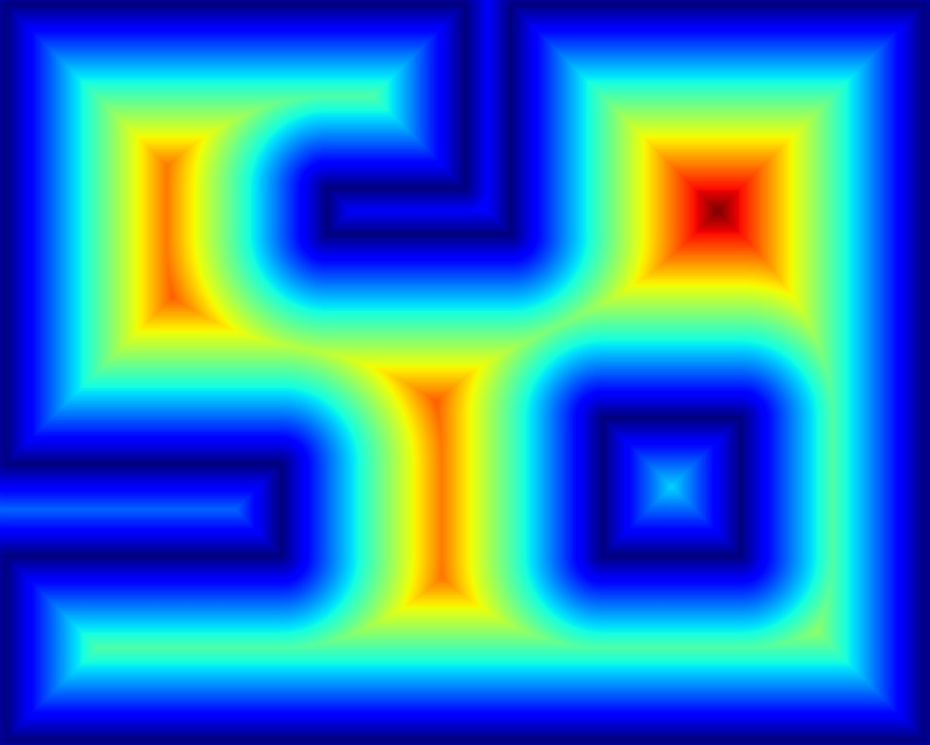} &
        
        \includegraphics[width=0.18\textwidth]{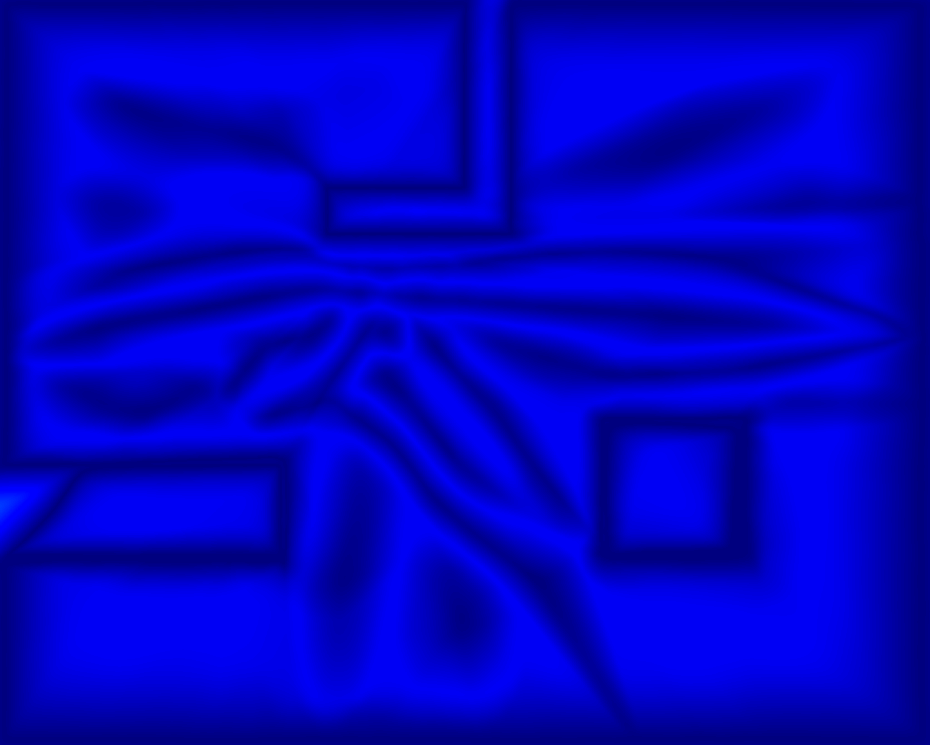} &
        \includegraphics[width=0.18\textwidth]{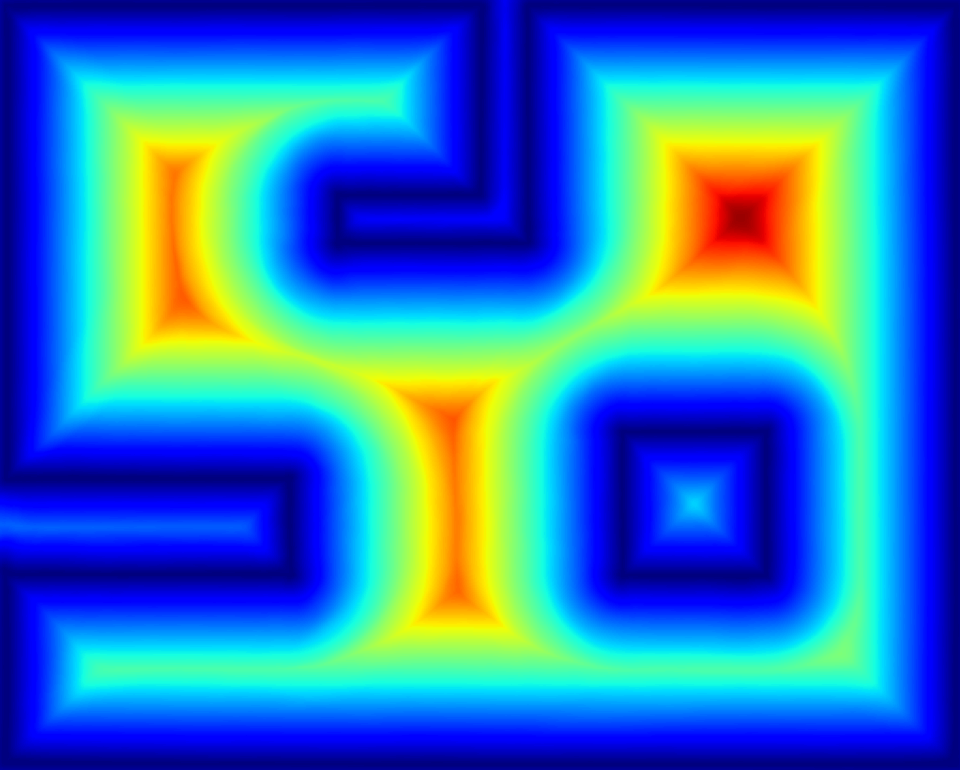} &
        \includegraphics[width=0.18\textwidth]{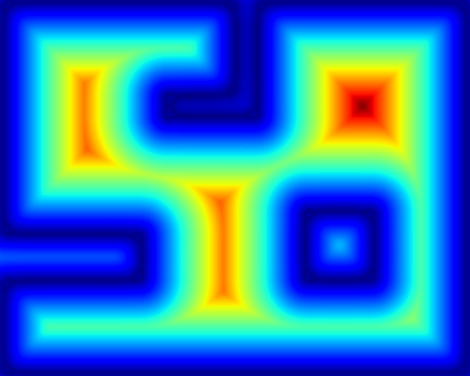} &
        \includegraphics[width=0.215\textwidth]{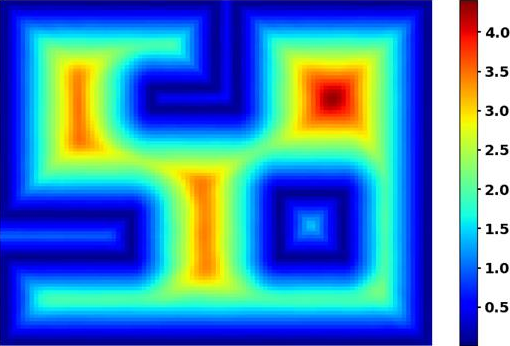} \\
        \scriptsize GT &  \scriptsize HotSpot & \scriptsize CAP-UDF & \scriptsize \ac{GPDF} &\scriptsize Ours \\[0.3em]

    \end{tabular}

    \caption{Qualitative comparison of 2D \ac{DF} predictions for the standalone formulation of the proposed method against GT, HotSpot, CAP-UDF, and \ac{GPDF} on the Snowflake (top row) and Gazebo (bottom row) datasets.}
     
    \label{fig:2d_df_benchmark}
\end{figure*}

\noindent\textbf{2D distance field and gradient evaluation.}
Figure~\ref{fig:2d_df_benchmark} and Table~\ref{tab:2d_df_benchmark} compare the standalone formulation against the \ac{DF}-only methods. Our method achieves the best RMSE on Gazebo ($0.02$) and remains close to the strongest benchmark on Snowflake. It also gives the most consistent gradient directions, matching the best mean cosine similarity with GPDF on Snowflake and achieving the highest value on Gazebo ($0.98$). While not best in gradient-norm MAE, its values remain comparable to the benchmarks.

\begin{table}[t]
    \centering
    \caption{2D \ac{DF} comparison across methods. Best results are shown in bold.}
    \label{tab:2d_df_benchmark}
    \resizebox{0.8\textwidth}{!}
    {
    \begin{tabular}{l|cccc|cccc}
        \toprule
        \multirow{2}{*}{Metric}
        & \multicolumn{4}{c|}{Snowflake ($1.2\,\text{m}$ x  $1.2\,\text{m}$)}
        & \multicolumn{4}{c}{Gazebo ($20\,\text{m}$ x  $16\,\text{m}$)} \\
        \cmidrule(lr){2-5} \cmidrule(lr){6-9}
        & HotSpot & CAP-UDF & \ac{GPDF} & Ours
        & HotSpot & CAP-UDF & \ac{GPDF} & Ours \\
        \midrule
        RMSE $\downarrow$
            & 0.014 & \textbf{0.004} & 0.008 & 0.005
            & 1.4 & 0.04 & 0.04 & \textbf{0.02} \\
        Mean cos. sim. of $\nabla D$ $\uparrow$
            & 0.94 & 0.96 & \textbf{0.94} & \textbf{0.94}
            & 0.42 & 0.97 & 0.93 & \textbf{0.98} \\
        MAE of $(\|\nabla D\|-1)$ $\downarrow$
            & 0.12 & 0.07 & \textbf{0.05} & 0.08
            & 0.65 & \textbf{0.07} & 0.08 & 0.09 \\
        \bottomrule
    \end{tabular}%
    }
\end{table}

\noindent\textbf{3D distance field and gradient evaluation.}
\begin{table}[th]
    \centering
    \caption{3D \ac{DF} comparison across methods. Best results are shown in bold.}
    \label{tab:3d_df_benchmark}
    \scriptsize
    \setlength{\tabcolsep}{2.2pt}
    \renewcommand{\arraystretch}{1.0}
    \resizebox{\columnwidth}{!}{%
    \begin{tabular}{lccccc|ccccc}
        \toprule
        \multirow{2}{*}{Metric}
        & \multicolumn{5}{c|}{TUM ($8.5\,\text{m}$ x  $6.8\,\text{m} $ x $1.8\,\text{m} $)}
        & \multicolumn{5}{c}{Cow\&Lady ($10.6\,\text{m}$ x  $11.9\,\text{m} $ x $3.4\,\text{m} $)} \\
        \cmidrule(lr){2-6} \cmidrule(lr){7-11}
        &  HotSpot & CAP-UDF & \ac{GPDF} & Standalone & Joint
        &  HotSpot & CAP-UDF & \ac{GPDF} & Standalone & Joint \\
        \midrule
        RMSE slice $\downarrow$
        & 0.80 & 0.43 & 0.04 & \textbf{0.016} & 0.019
        &  0.90 & 0.20 &0.05 & \textbf{0.03} & \textbf{0.03} \\

        RMSE full $\downarrow$
        & 0.96 & 0.60 &0.046 &  \textbf{0.020} & 0.024
        &  1.8 & 0.30 & 0.08 &\textbf{0.035} & 0.042 \\

        Mean cos. sim. of $\nabla D$ $\uparrow$
        &  0.14 & 0.87 & 0.93 &0.90 & \textbf{0.96}
        &  0.24 & 0.69 & 0.82 & 0.84 & \textbf{0.89} \\

        MAE$(\|\nabla D\|-1)$ $\downarrow$
        & 0.80 & 0.84 &\textbf{0.04}  &  \textbf{0.04} & \textbf{0.04}
        & 0.91 & 0.29 &  0.16 & \textbf{0.14} & 0.15 \\
        \bottomrule
    \end{tabular}%
    }
\end{table}
Table~\ref{tab:3d_df_benchmark} compares the proposed standalone and joint formulations against the \ac{DF}-only methods. Ours achieves the best distance accuracy with the standalone model, giving the lowest full-volume RMSE on TUM ($0.020$) and Cow\&Lady ($0.035$). The joint model remains close in RMSE and gives the best gradient direction, achieving the highest mean cosine similarity in both datasets. Gradient-norm MAE is also competitive, matching \ac{GPDF} on TUM ($0.04$) and giving the lowest value on Cow\&Lady with the standalone model ($0.14$). Qualitative results in Figure~\ref{fig:3d_df_benchmark} show the same trend, with the proposed formulations closely matching the ground-truth \ac{DF} across both near-surface and free-space regions.

Lastly, we report the training and average per point query times taken across datasets for \ac{DF} modelling (Table~\ref{tab:runtime_comparison}). Ours, apart from Snowflake, achieves significantly lower training time among the methods. Furthermore, our method's average query time per point scales much more favourably across datasets, particularly in 3D, second only to CAP-UDF.

\begin{figure}[t]
    \centering
    \setlength{\tabcolsep}{1pt}

    \begin{tabular}{cccccc}
        
\includegraphics[width=0.15\columnwidth]{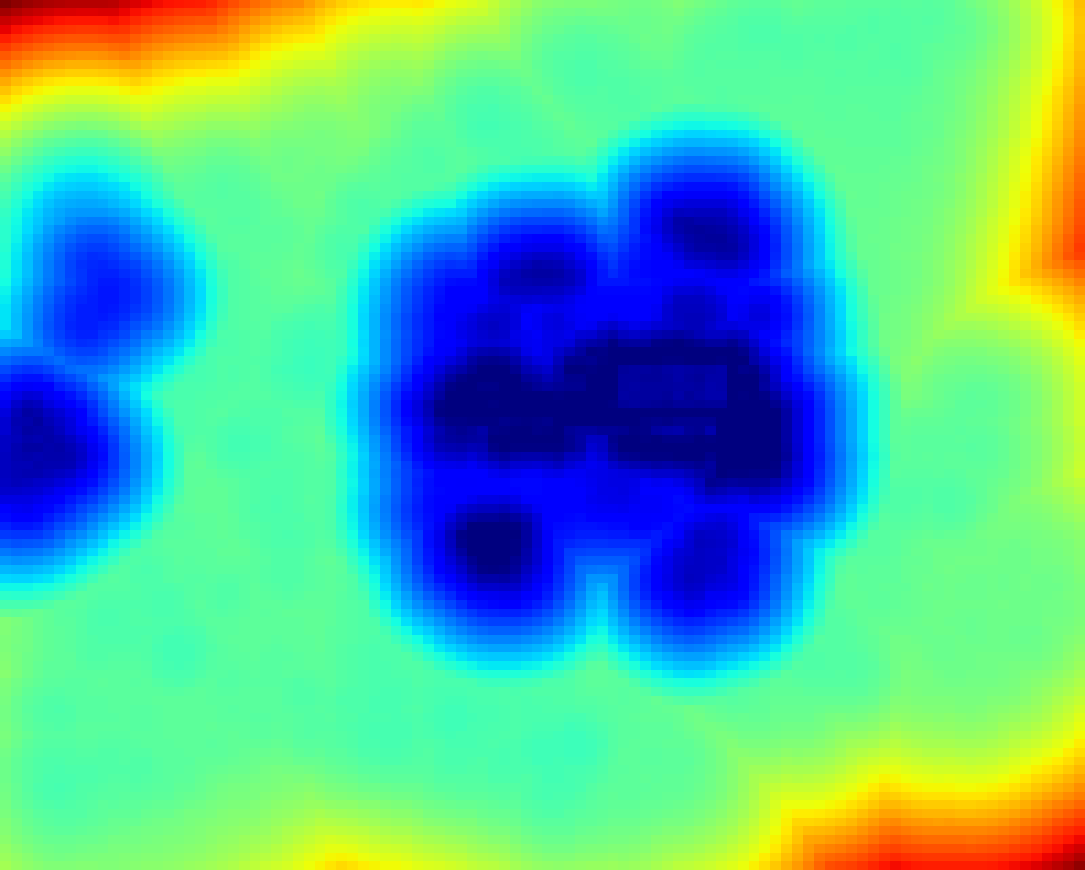} &
       
        \includegraphics[width=0.15\columnwidth]{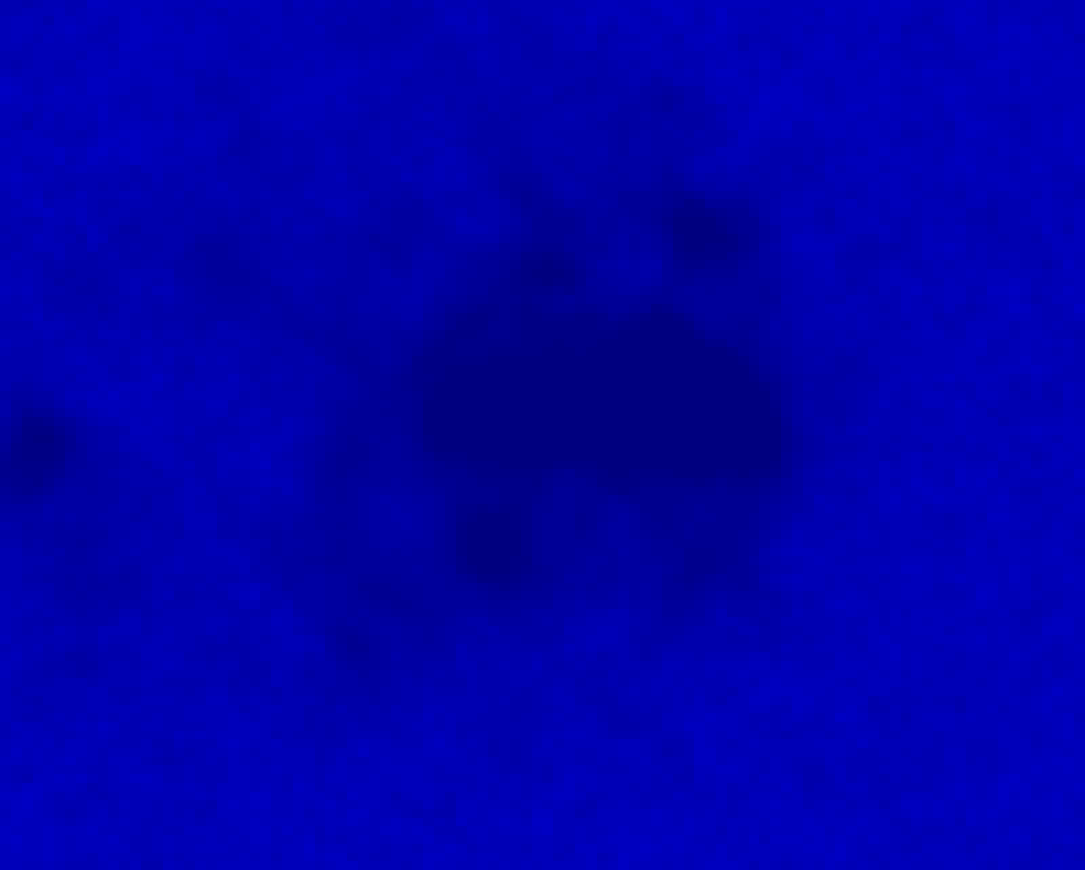} &
        \includegraphics[width=0.15\columnwidth]{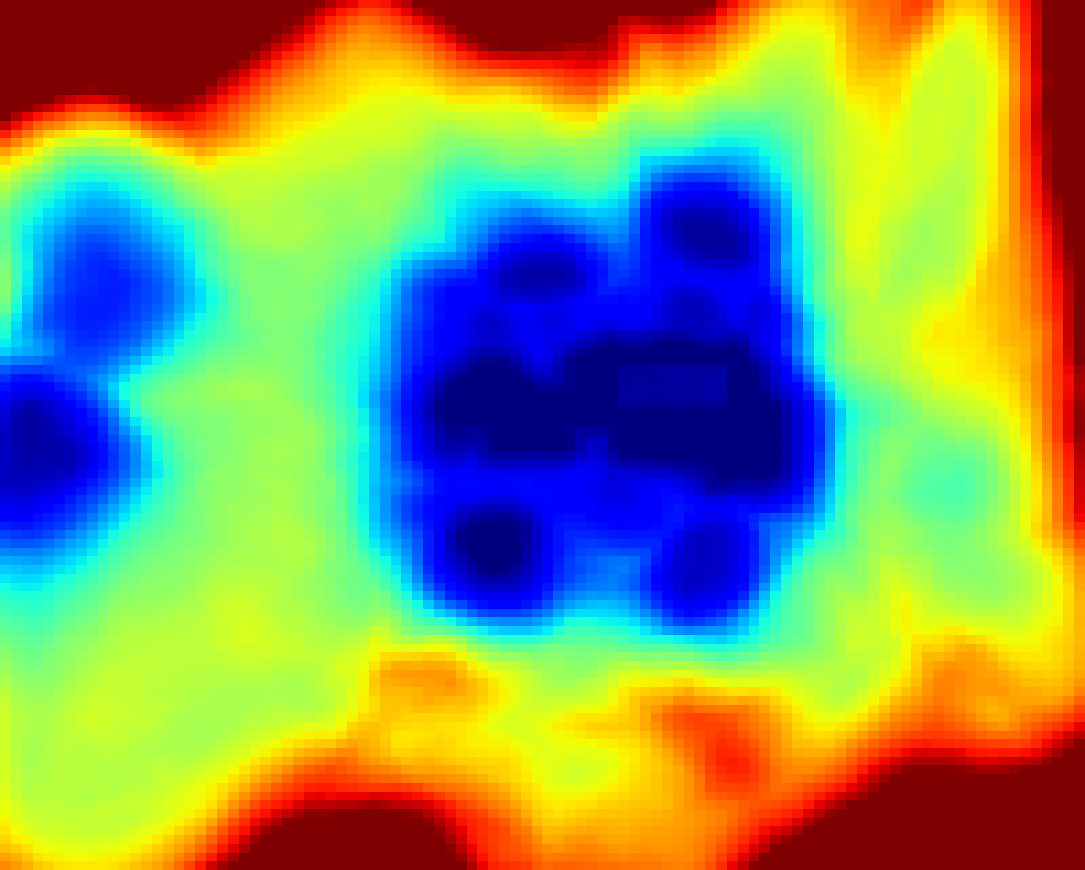} &
         \includegraphics[width=0.15\columnwidth]{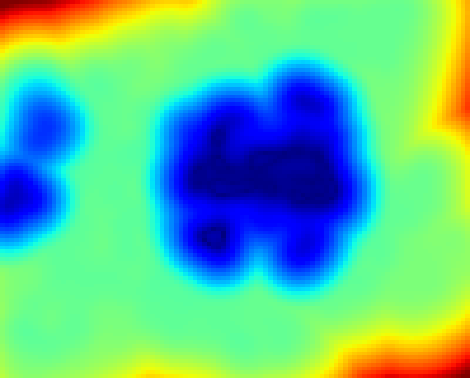} &
        \includegraphics[width=0.15\columnwidth]{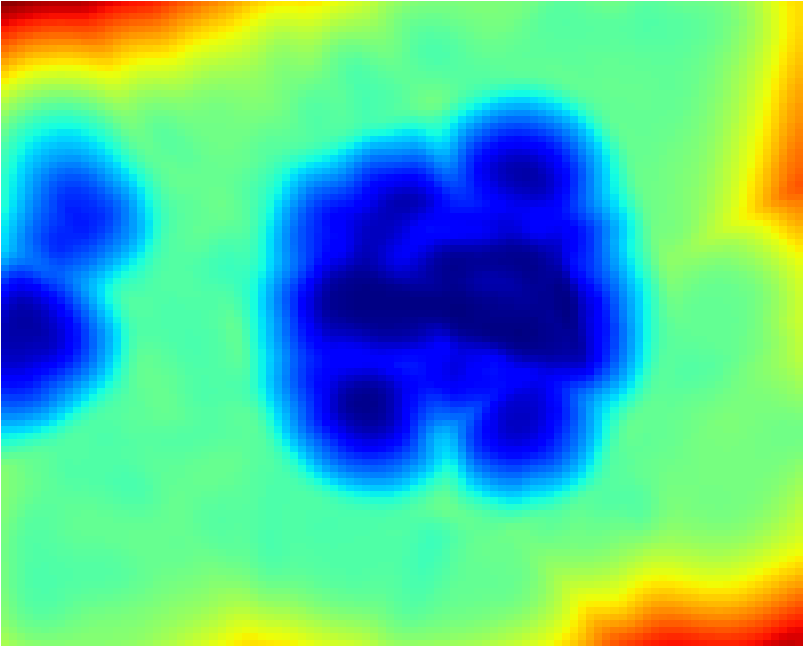} &
        \includegraphics[width=0.18\columnwidth] {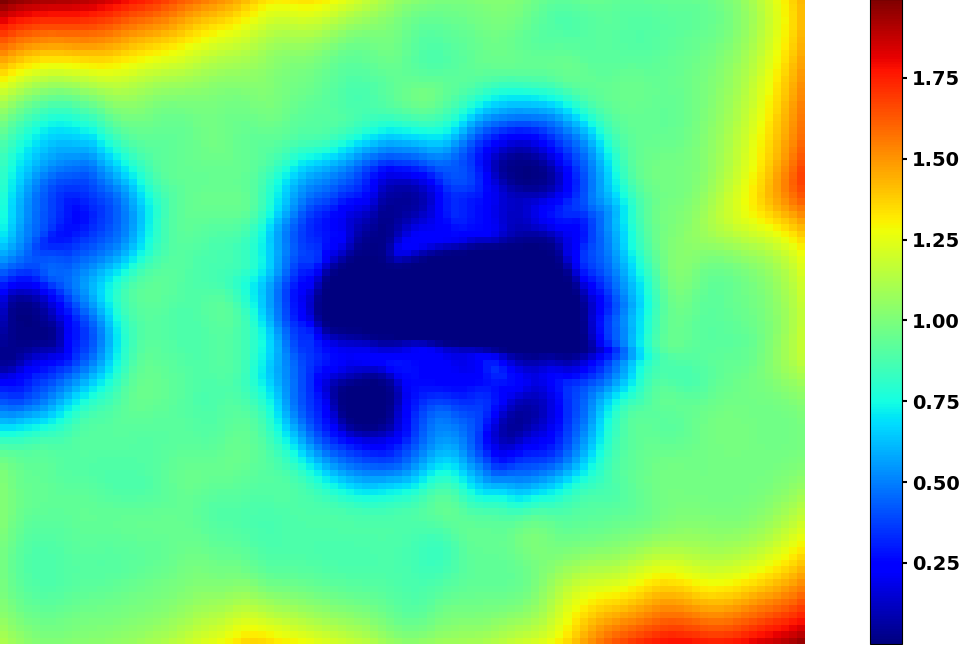}  \\
        \includegraphics[width=0.15\columnwidth]{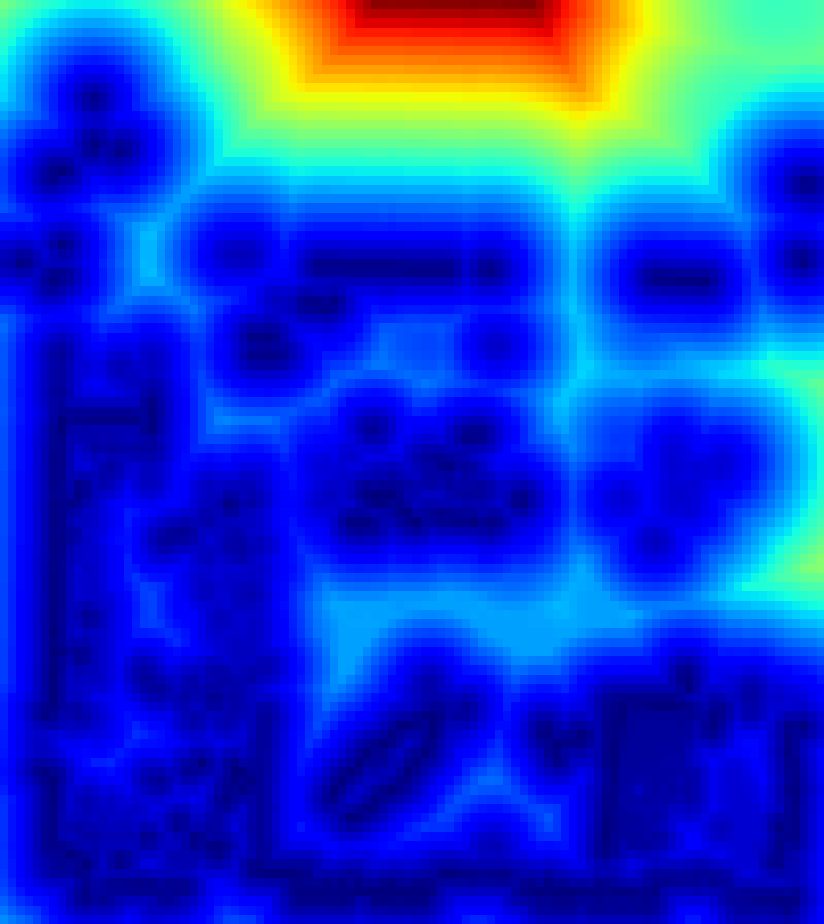} &
        
        \includegraphics[width=0.15\columnwidth]{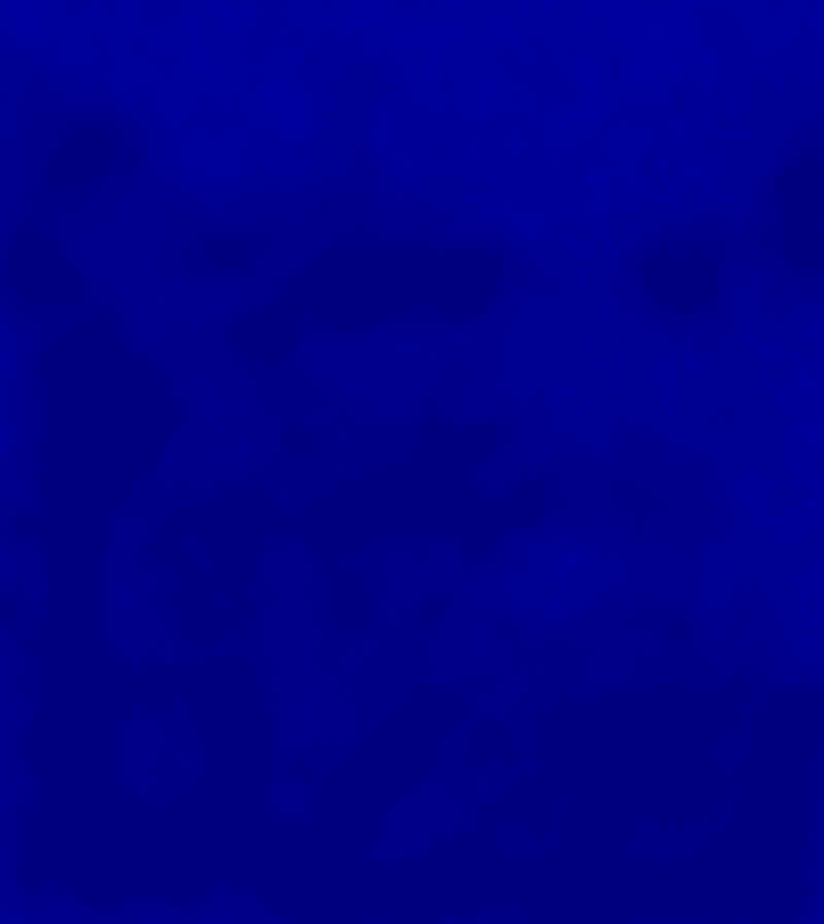} &
        \includegraphics[width=0.15\columnwidth]{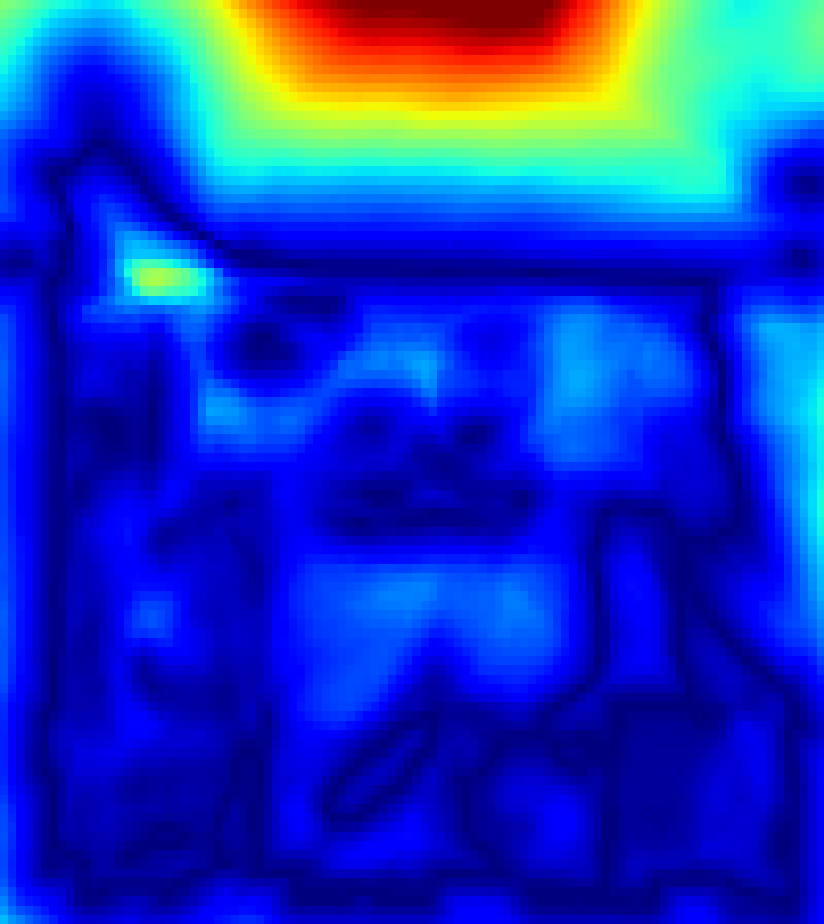} &
        \includegraphics[width=0.15\columnwidth]{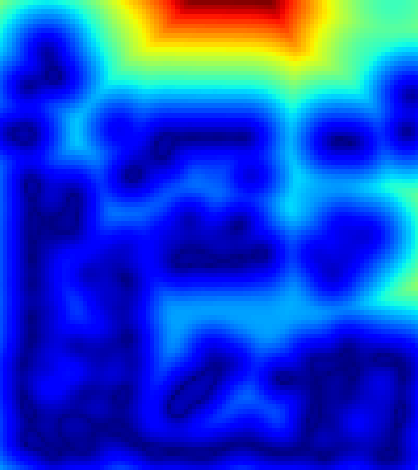} &
        \includegraphics[width=0.15\columnwidth]{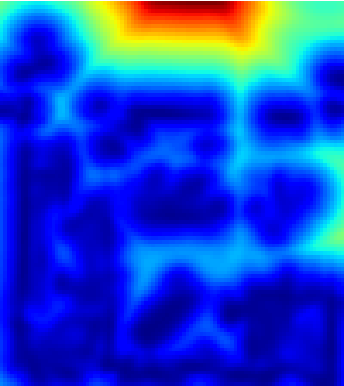} &
        \includegraphics[width=0.18\columnwidth] {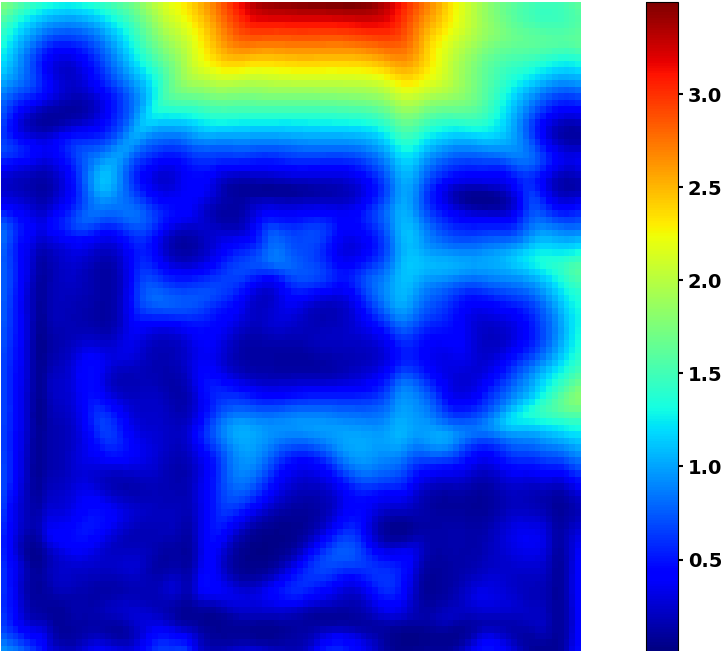} \\
     
    \scriptsize GT &  \scriptsize HotSpot & \scriptsize CAP-UDF & \scriptsize \ac{GPDF} & \scriptsize Ours (SA) & \scriptsize Ours (Joint) \\[0.2em]
    \end{tabular}

     \caption{Qualitative comparison of 3D \ac{DF} predictions on a slice (z=1.0) of the TUM (top) and Cow\&Lady (bottom) dataset. From left to right: GT, HotSpot, CAP-UDF, \ac{GPDF}, the proposed standalone, and the joint formulations.}
    \label{fig:3d_df_benchmark}
\end{figure}

\begin{table*}[t]
    \centering
    \caption{Runtime comparison among \ac{DF}-only methods across datasets: training time (s) and average query time per point ($\mu$s/point).}
    \label{tab:runtime_comparison}
    \scriptsize
    \setlength{\tabcolsep}{2.4pt}
    \renewcommand{\arraystretch}{0.9}
    \resizebox{\textwidth}{!}{%
    \begin{tabular}{lcc|cc|cc|cc}
        \toprule
        \multirow{2}{*}{Method}
        & \multicolumn{2}{c|}{Snowflake}
        & \multicolumn{2}{c|}{Gazebo}
        & \multicolumn{2}{c|}{TUM}
        & \multicolumn{2}{c}{Cow\&Lady} \\
        \cmidrule(lr){2-3}
        \cmidrule(lr){4-5}
        \cmidrule(lr){6-7}
        \cmidrule(lr){8-9}
        & Train  $\downarrow$ & Query/pt  $\downarrow$
        & Train  $\downarrow$ & Query/pt  $\downarrow$
        & Train  $\downarrow$ & Query/pt  $\downarrow$
        & Train  $\downarrow$ & Query/pt  $\downarrow$ \\
        \midrule
        HotSpot
        & 152 & \textbf{0.03} & 174 & \textbf{0.04} & 5333 & 203.4 & 5251 & 198.6  \\
        CAP-UDF
        & 405 & 0.07 & 427 & 0.08 & 1166 & \textbf{0.26} & 1361 & \textbf{0.24} \\
        \ac{GPDF}
        &  \textbf{7.5}& 1.28 & 126.5& 5.95 & 624.3 & 100.88 & 492.8 & 89.97 \\
        
        Ours (SA)
        & 32 & \textbf{0.03}  & \textbf{74} & 0.06 &  \textbf{151} &  0.45 & \textbf{182.6} & 0.52 \\
        \bottomrule
    \end{tabular}%
    }
    \vspace{-1.53em}
\end{table*}
\subsection{Rendering Performance}
Both vanilla \ac{2DGS} and the proposed joint formulation are trained for $30$k iterations and initialised from the same input point cloud. For rendering, both methods use the same \ac{2DGS}'s normal and depth-distortion regularisation weights $\lambda_{\mathrm{norm}}=0.05$, and $\lambda_{\mathrm{depth}}=0.1$. For the joint formulation, we set the centre-consistency and disk-consistency losses weighted by $0.5$ each, and use $K_c$ = 5000 disks and $N_s$ = 1000 points per disk in each iteration for the coupling loss calculation.

\begin{figure*}[t]
    \centering
    \begin{subfigure}[b]{0.32\textwidth}
        \includegraphics[width=\linewidth, height=0.7\columnwidth]{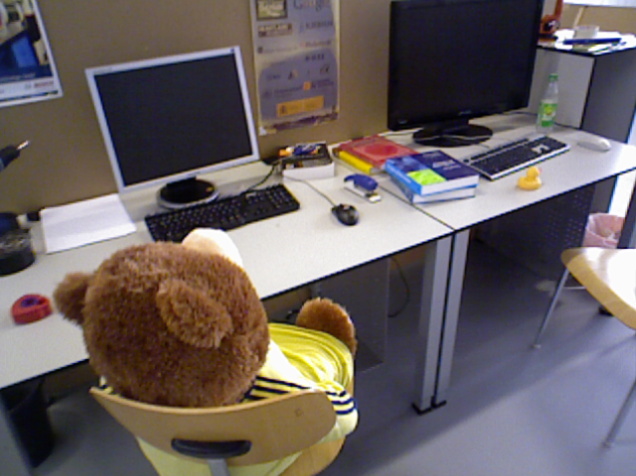}\\[0.2em]
        \includegraphics[width=\linewidth, height=0.7\columnwidth]{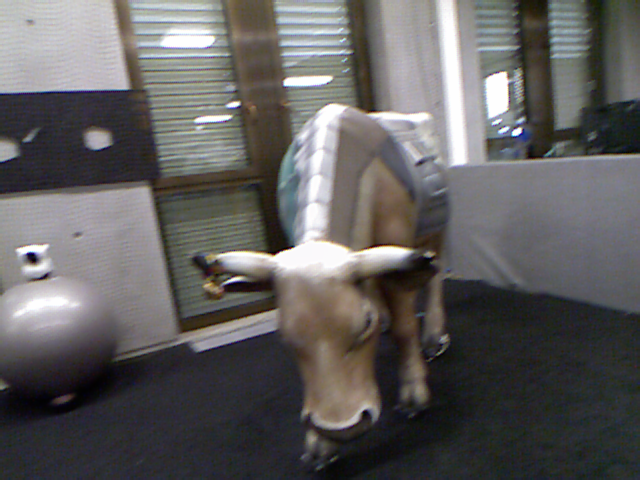}
        \caption{GT}
        \label{fig:render_gt}
    \end{subfigure}\hfill
    \begin{subfigure}[b]{0.32\textwidth}
        \includegraphics[width=\linewidth, height=0.7\columnwidth]{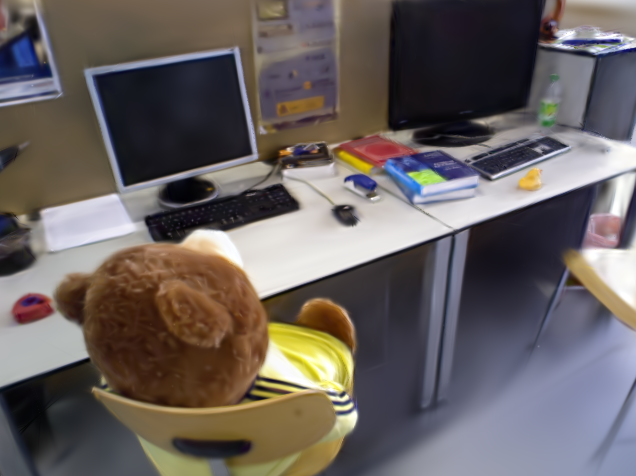}\\[0.2em]
        \includegraphics[width=\linewidth, height=0.7\columnwidth]{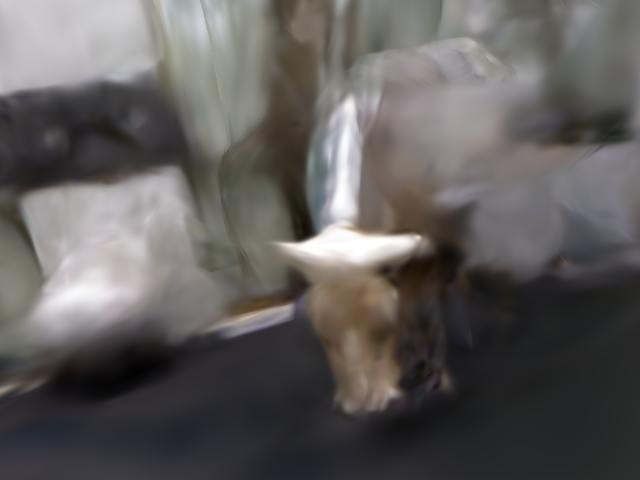}
        \caption{\ac{2DGS}}
        \label{fig:render_2dgs}
    \end{subfigure}\hfill
    \begin{subfigure}[b]{0.32\textwidth}
        \includegraphics[width=\linewidth, height=0.7\columnwidth]{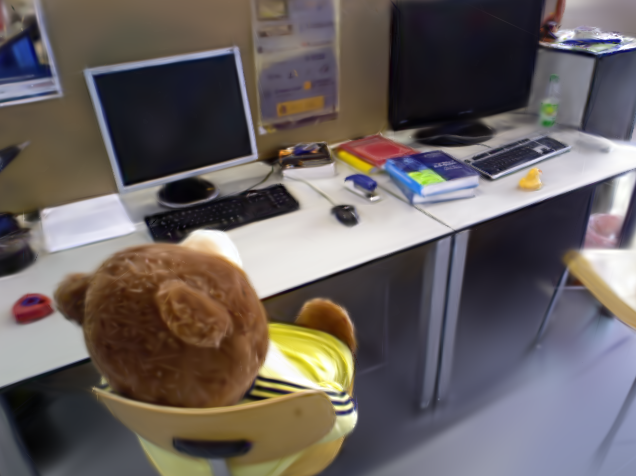}\\[0.2em]
        \includegraphics[width=\linewidth, height=0.7\columnwidth]{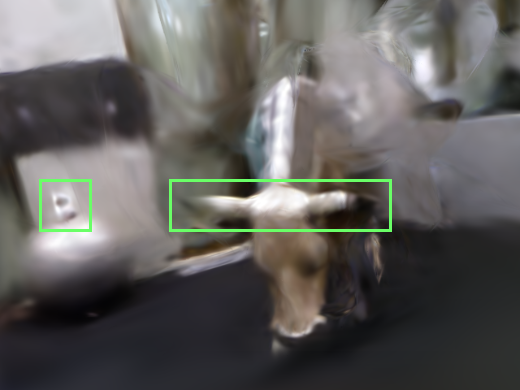}
        \caption{Ours (Joint)}
        \label{fig:render_ours}
    \end{subfigure}
    \caption{Qualitative comparison of rendering results on the TUM (top row) and Cow\&Lady (bottom row) dataset.}
    \label{fig:rendering}
\end{figure*}

Figure~\ref{fig:rendering} shows qualitative novel-view rendering results, and Table~\ref{tab:rendering} reports quantitative metrics over the full train-test image set. 
Overall, the joint formulation improves most rendering metrics over vanilla \ac{2DGS}. 
On TUM, it improves all reported appearance metrics. On Cow\&Lady, it improves PSNR by $1.09$ dB and reduces LPIPS and L1 error, while SSIM remains comparable. The qualitative results show that the joint formulation preserves the main scene structure and produces renderings that are visually comparable on TUM and improved on Cow\&Lady, as highlighted in Figure~\ref{fig:render_ours}.

\begin{table}[t]
    \centering
    \caption{Rendering comparison across datasets. Best results are shown in bold.}
    \label{tab:rendering}
    \scriptsize
    \setlength{\tabcolsep}{1.6pt}
    \renewcommand{\arraystretch}{1.0}
    \begin{adjustbox}{max width=\columnwidth}
    \begin{tabular}{lcccc|cccc}
        \toprule
        \multirow{2}{*}{Method}
        & \multicolumn{4}{c|}{TUM}
        & \multicolumn{4}{c}{Cow\&Lady} \\
        \cmidrule(lr){2-5} \cmidrule(lr){6-9}
        & PSNR $\uparrow$ & SSIM $\uparrow$ & LPIPS $\downarrow$ & L1 $\downarrow$
        & PSNR $\uparrow$ & SSIM $\uparrow$ & LPIPS $\downarrow$ & L1 $\downarrow$ \\
        \midrule
        \ac{2DGS}
        & 27.54 & 0.909 & 0.174 & 0.038
        & 18.84 & \textbf{0.638} & 0.603 & 0.080 \\

        Ours
        & \textbf{28.13} & \textbf{0.922} & \textbf{0.160} & \textbf{0.027}
        & \textbf{19.93} & 0.633 & \textbf{0.515} & \textbf{0.073} \\
        \bottomrule
    \end{tabular}
    \end{adjustbox}
\end{table}

\subsection{Surface Reconstruction Results}
\begin{figure}[t]
    \centering
    \setlength{\tabcolsep}{1pt}

    \begin{tabular}{cccccc}

        \includegraphics[width=0.16\columnwidth,height=0.12\columnwidth]{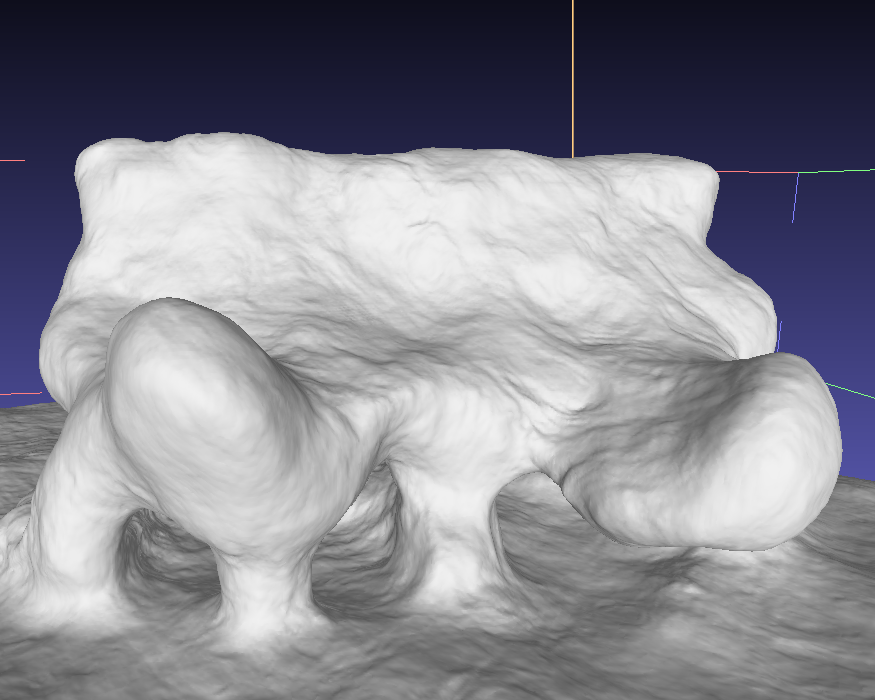} &
        \includegraphics[width=0.16\columnwidth,height=0.12\columnwidth]{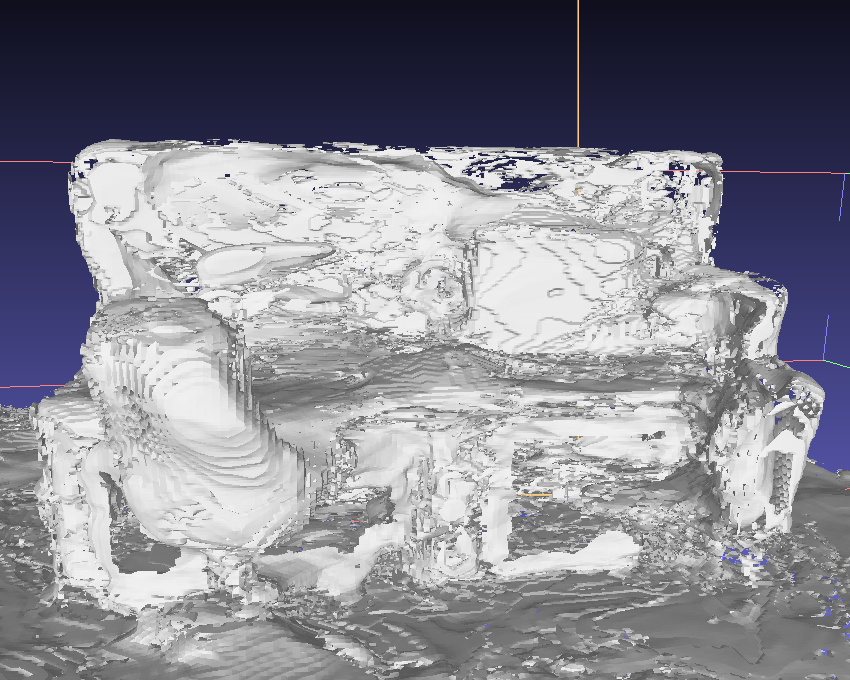} &
        \includegraphics[width=0.16\columnwidth,height=0.12\columnwidth]{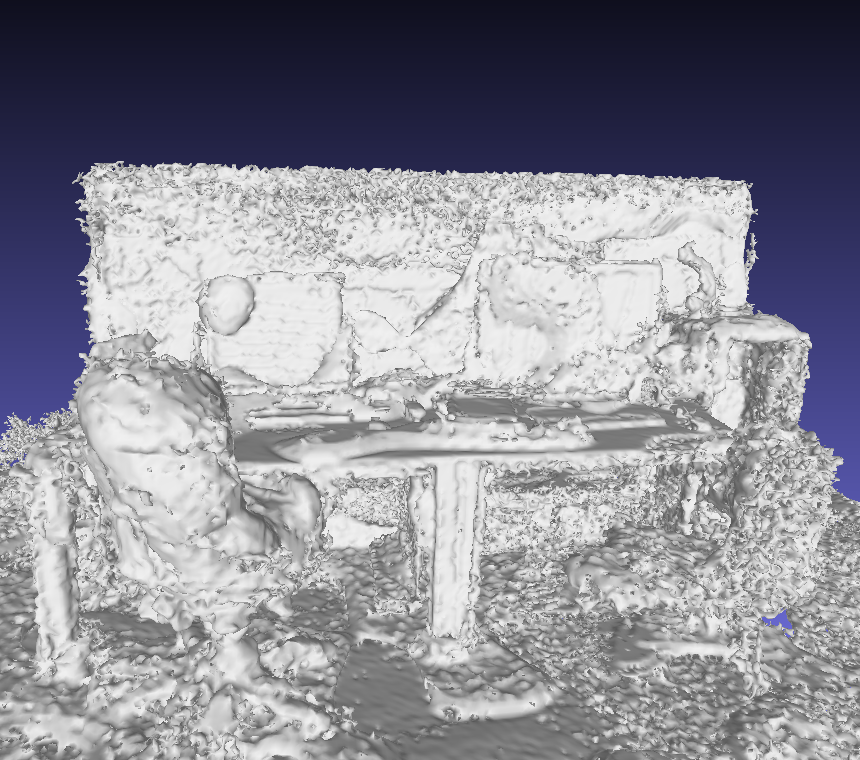} &
        \includegraphics[width=0.16\columnwidth,height=0.12\columnwidth]{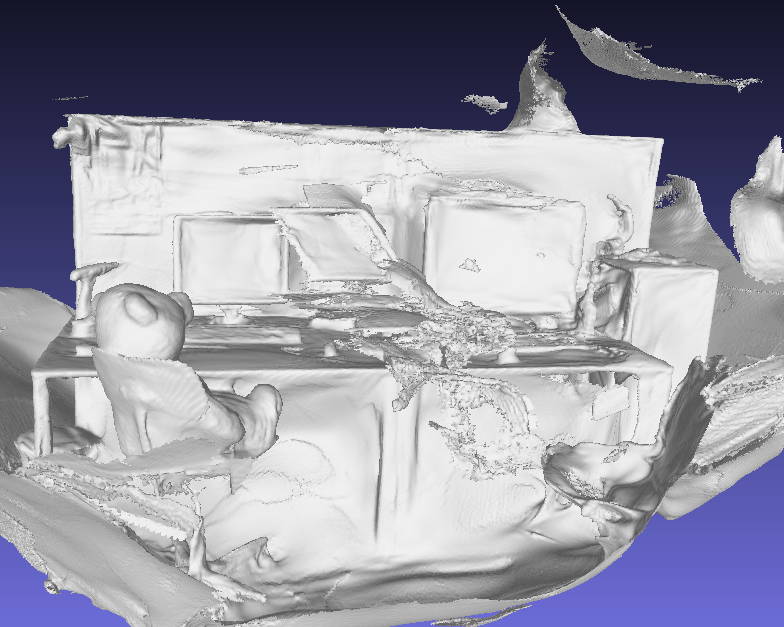} &
        \includegraphics[width=0.16\columnwidth,height=0.12\columnwidth]{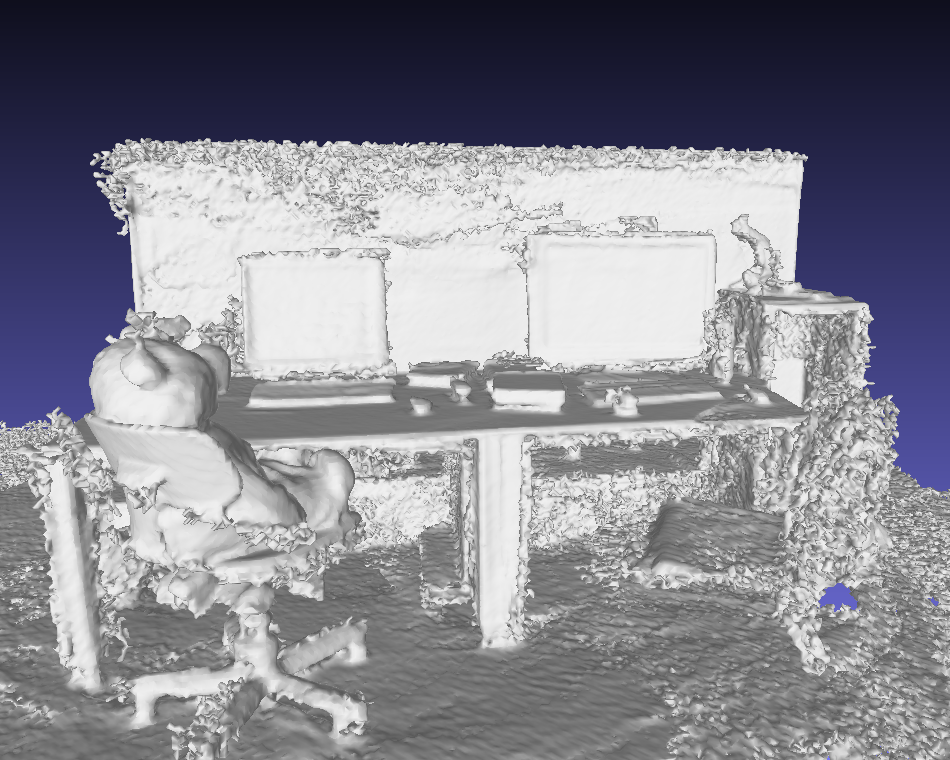} &
         \includegraphics[width=0.16\columnwidth,height=0.12\columnwidth]{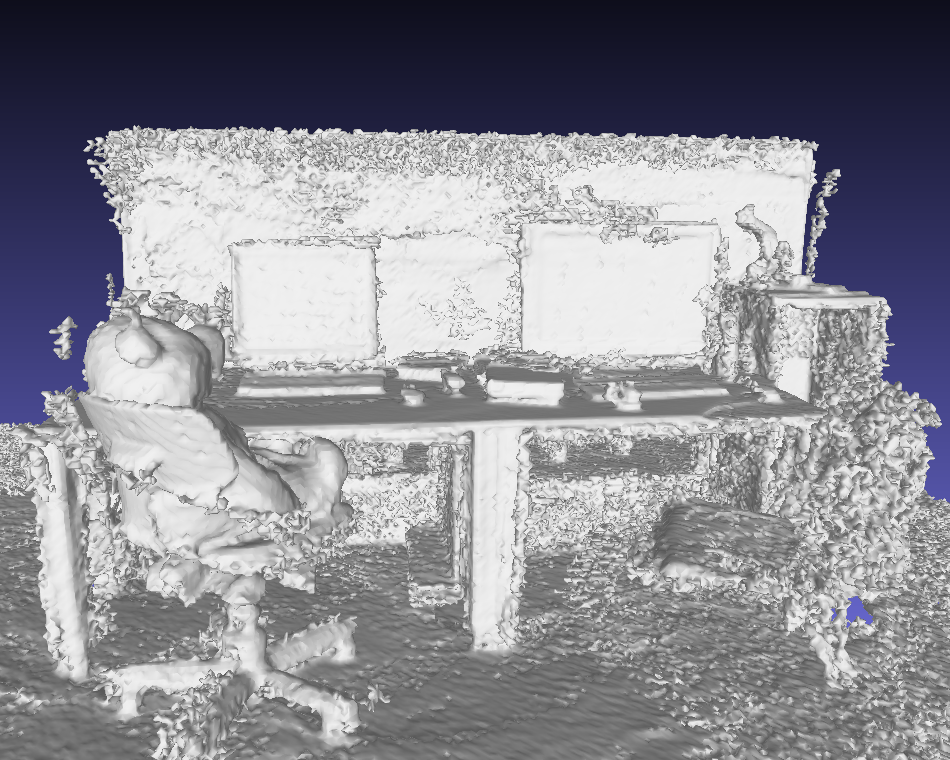} \\

        \includegraphics[width=0.16\columnwidth,height=0.12\columnwidth]{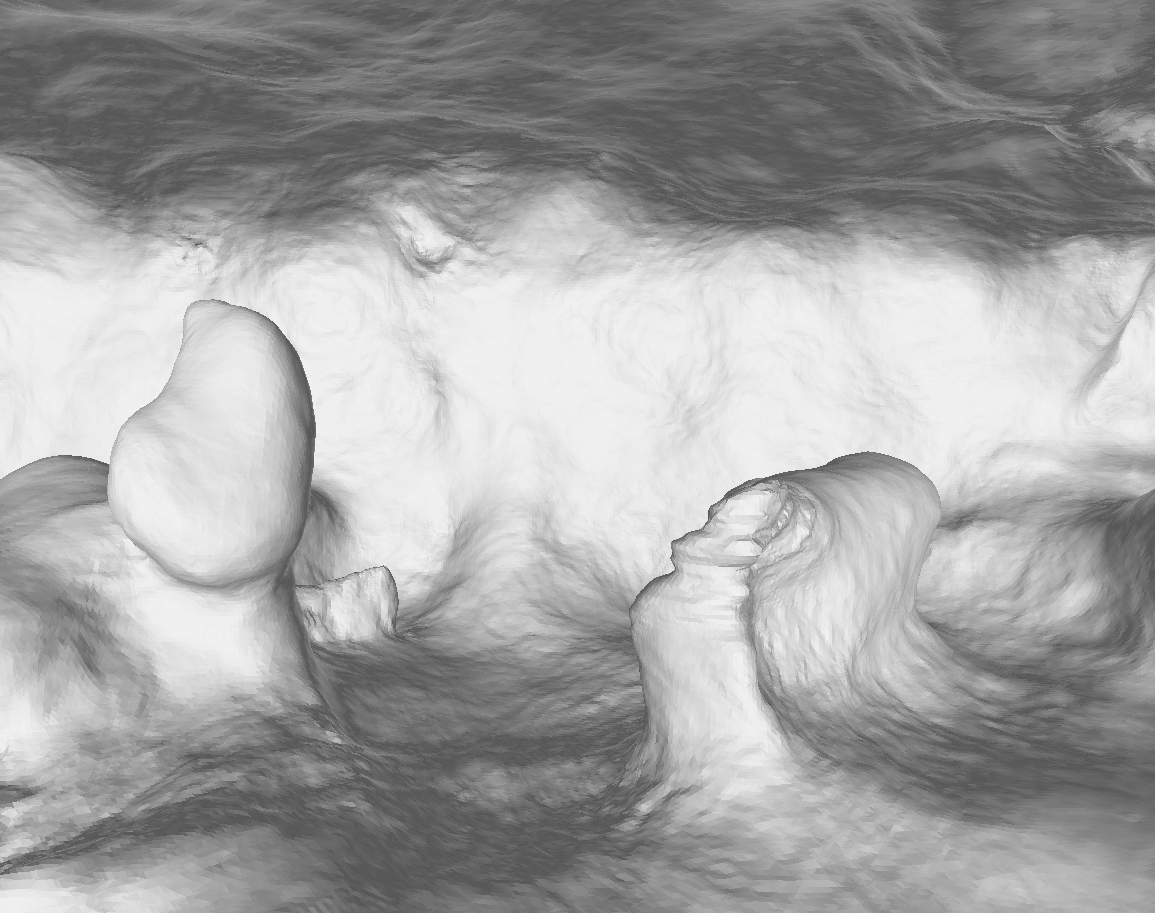} &
        \includegraphics[width=0.16\columnwidth,height=0.12\columnwidth]{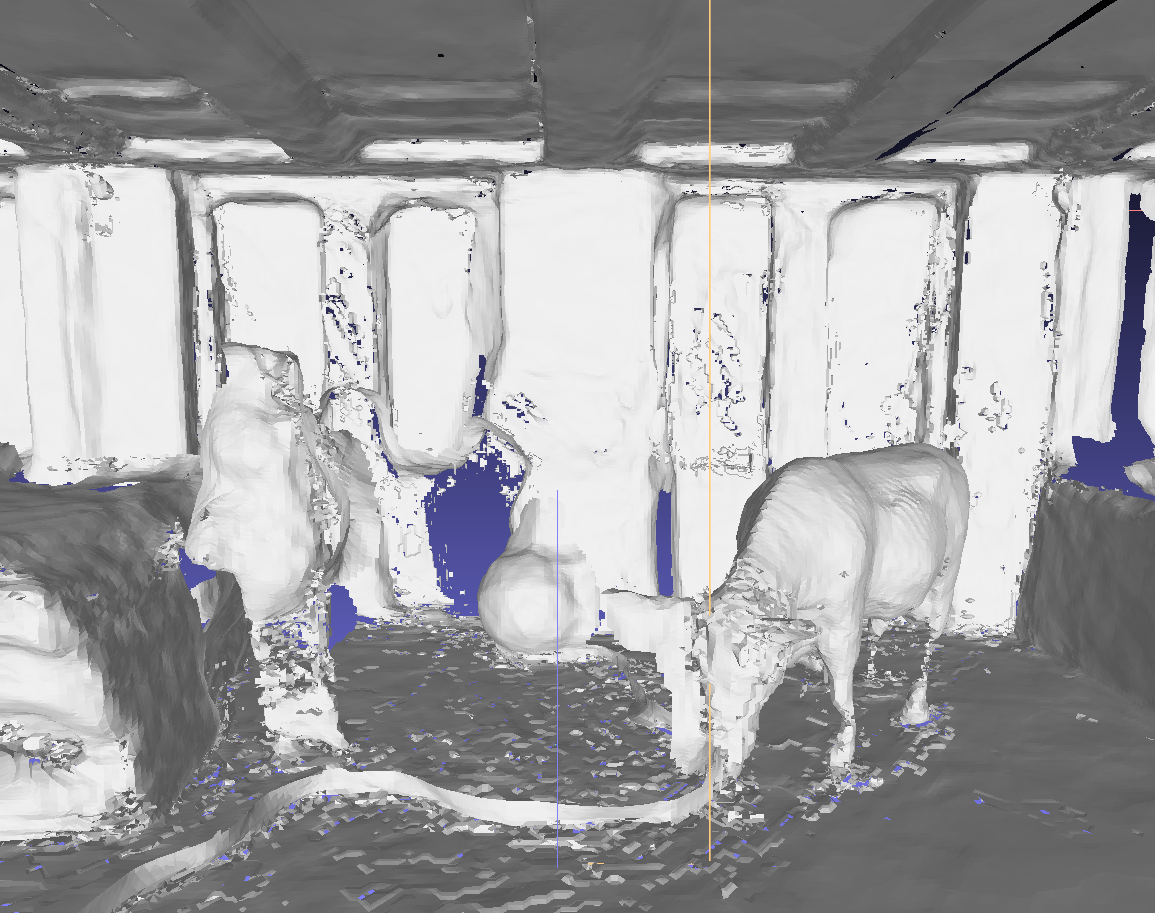} &
        \includegraphics[width=0.16\columnwidth,height=0.12\columnwidth]{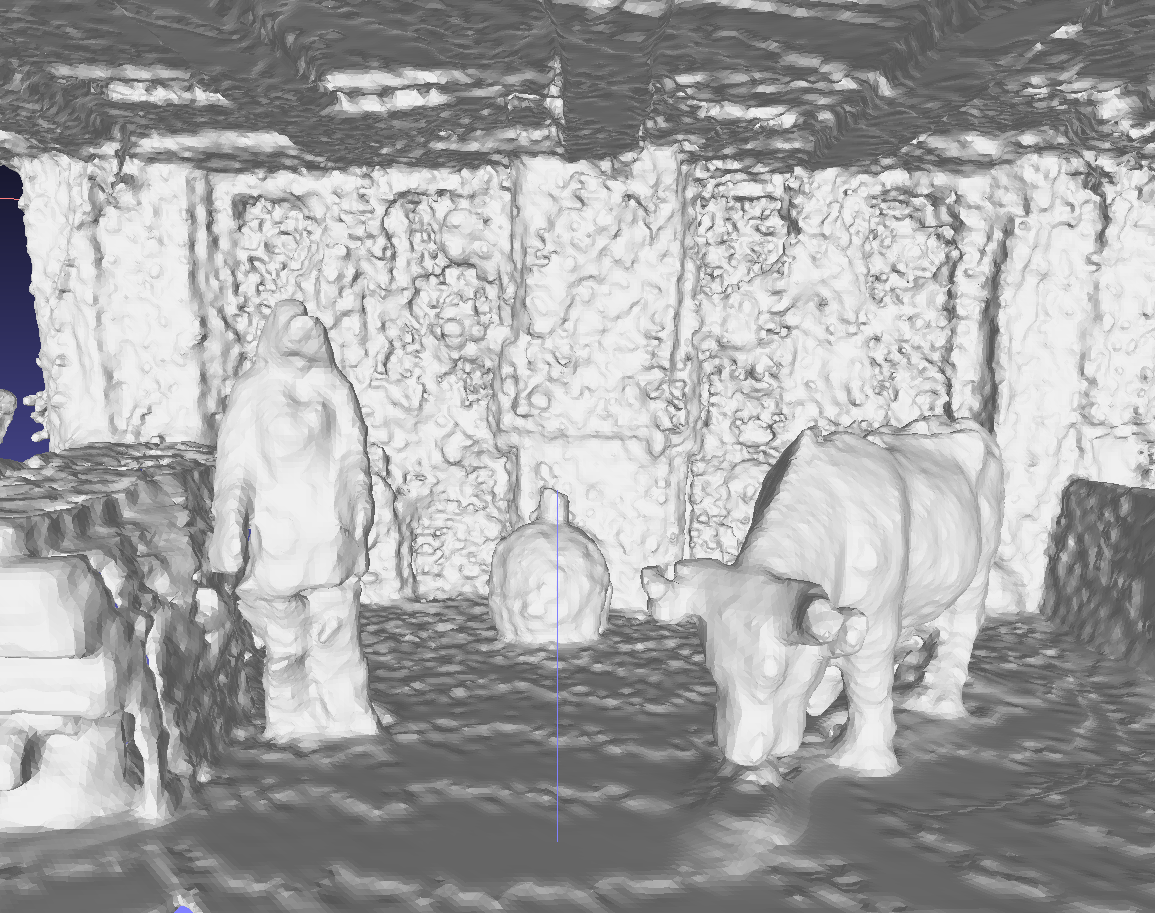} &
        \includegraphics[width=0.16\columnwidth,height=0.12\columnwidth]{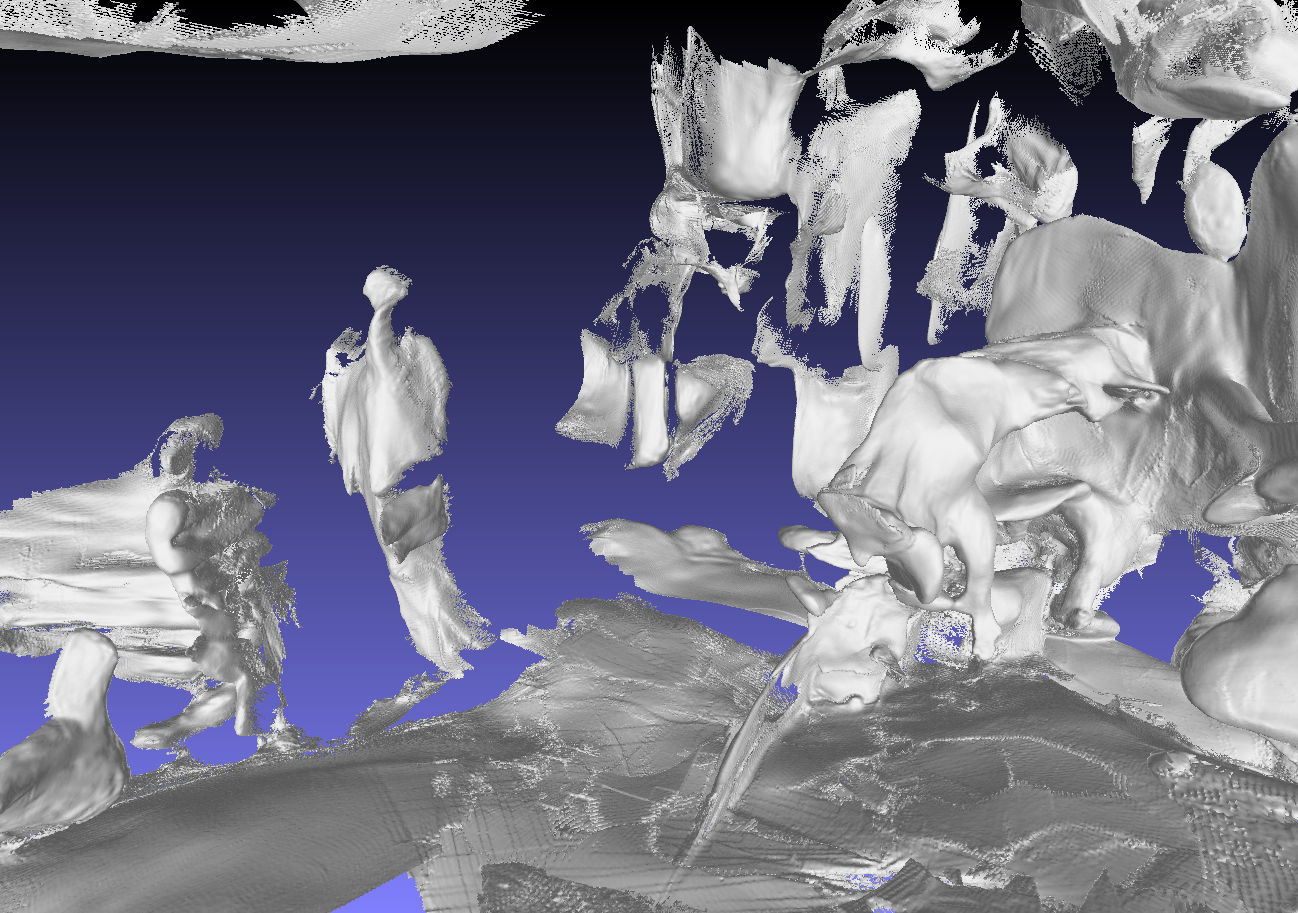} &
        \includegraphics[width=0.16\columnwidth,height=0.12\columnwidth]{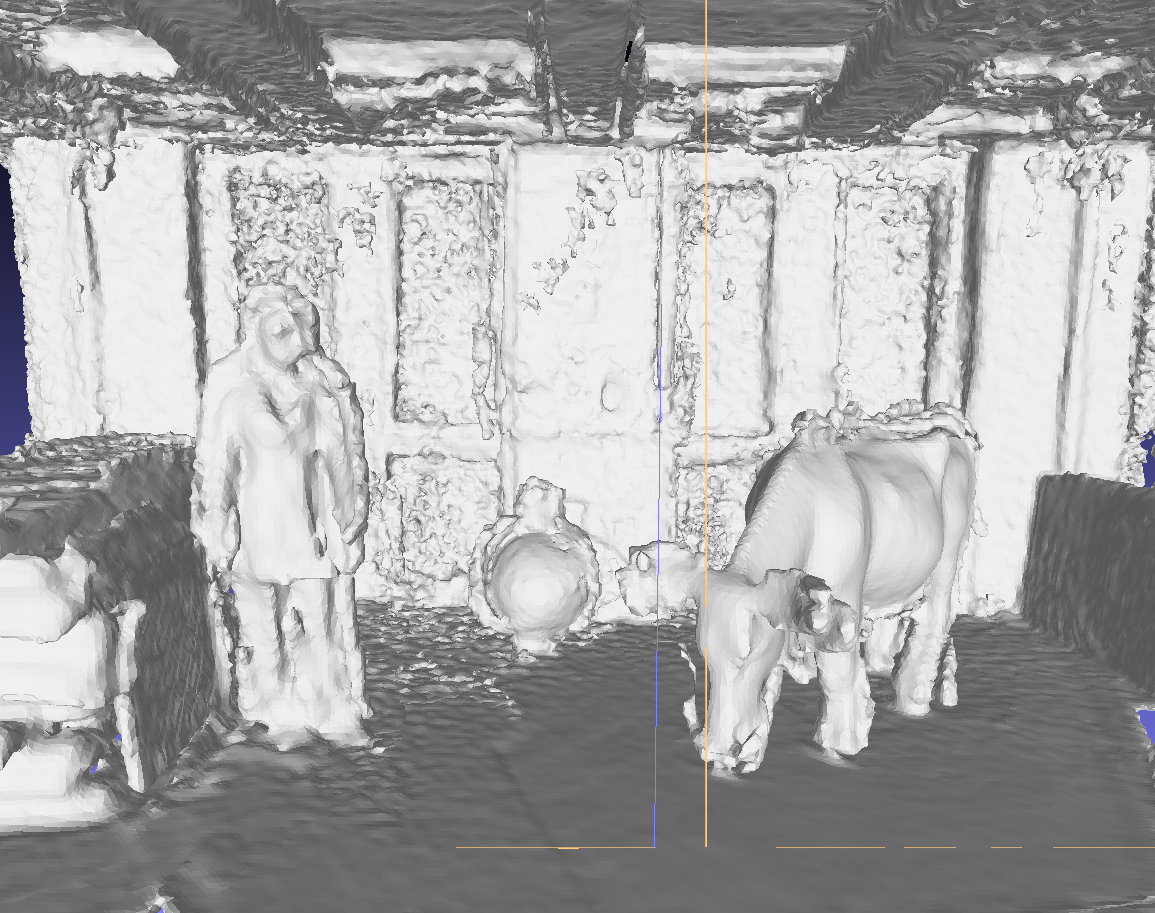} &
         \includegraphics[width=0.16\columnwidth,height=0.12\columnwidth]{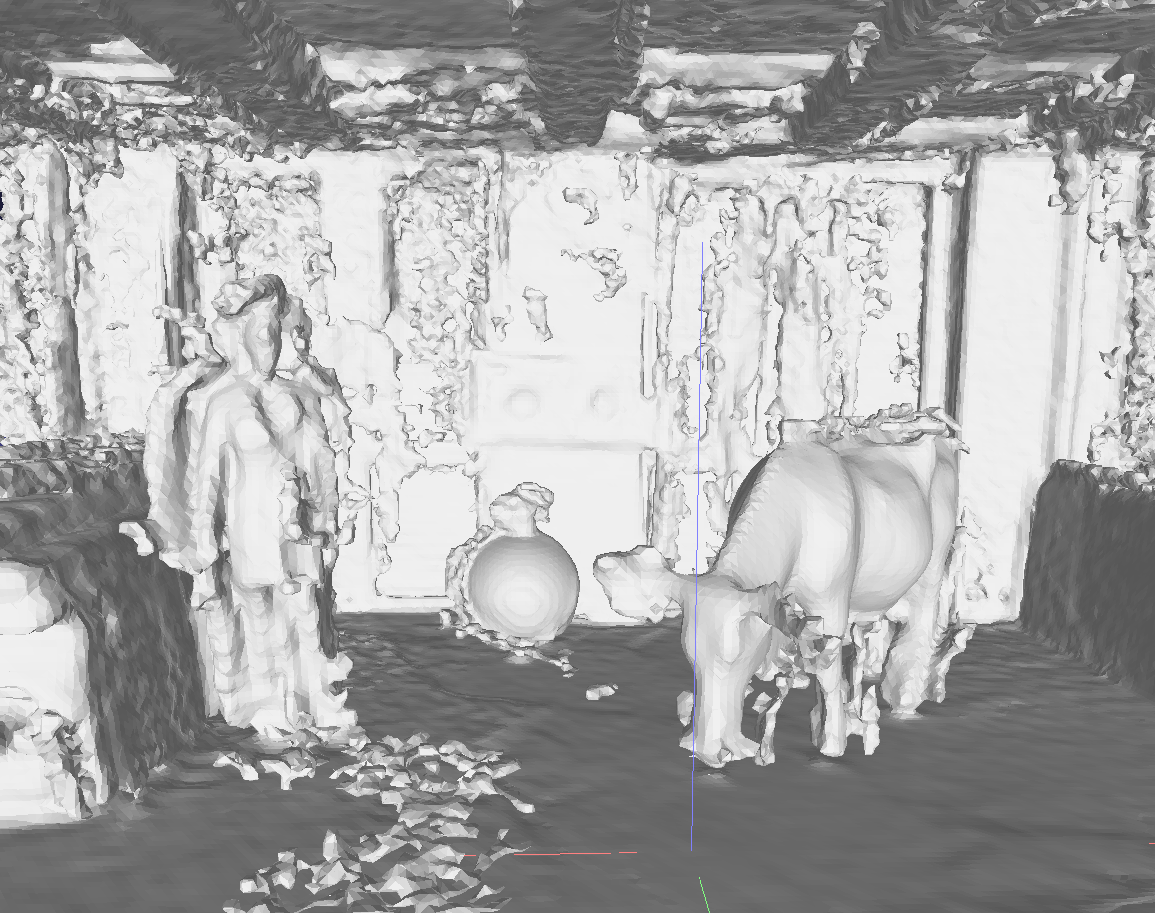} \\
     
      \scriptsize HotSpot & \scriptsize CAP-UDF & \scriptsize \ac{GPDF} & \scriptsize \ac{2DGS} & \scriptsize Ours (SA) &  \scriptsize Ours (Joint) \\[0.2em]
    \end{tabular}

    \caption{Qualitative comparison of mesh reconstruction of the TUM (top row) and Cow\&Lady (bottom row) dataset. From left to right: HotSpot, CAP-UDF, \ac{GPDF}, \ac{2DGS}, and Ours with the standalone and joint variants, respectively.}
    \label{fig:mesh_comparison}
\end{figure}

\begin{table}[t]
    \centering
    \caption{Surface reconstruction results across methods; best results in bold.}
    \label{tab:mesh_benchmark}
    \scriptsize
    \setlength{\tabcolsep}{2.2pt}
    \renewcommand{\arraystretch}{1.05}
    \resizebox{\columnwidth}{!}{%
    \begin{tabular}{lcccccc|cccccc}
        \toprule
        \multirow{2}{*}{Metric}
        & \multicolumn{6}{c|}{TUM}
        & \multicolumn{6}{c}{Cow\&Lady} \\
        \cmidrule(lr){2-7} \cmidrule(lr){8-13}
        &  HotSpot & CAP-UDF & \ac{GPDF} & \ac{2DGS} & Standalone & Joint
        &   HotSpot & CAP-UDF & \ac{GPDF} & \ac{2DGS} & Standalone & Joint \\
        \midrule
        Chamfer-L1  $\downarrow$
        &  0.021 & 0.021 & \textbf{0.016} & 0.41 & 0.020 & 0.023
        &  \textbf{0.032} & 0.044 & 0.056& 0.48 & 0.042 & 0.055\\
        \bottomrule
    \end{tabular}%
            }
\end{table}

Figure~\ref{fig:mesh_comparison} shows qualitative mesh reconstructions, and Table~\ref{tab:mesh_benchmark} reports mean Chamfer-L1 distances. For our standalone and joint formulations, meshes are extracted by forming a signed field from the predicted \acp{DF} and camera-ray fusion of posed depth frames, followed by marching cubes~\cite{marching}. 
The joint formulation can generate depth maps rendered from the jointly optimised \ac{2DGS} and use those to obtain a surface mesh similar to vanilla \ac{2DGS}'s own mesh reconstruction. However, we use the same \ac{DF}-based reconstruction for both proposed variants, as well as for \ac{GPDF}, to ensure direct comparability. Other methods' meshes are obtained from their publicly released codes. Errors are computed against the provided \ac{GT} point cloud for Cow\&Lady and a \ac{GT}-depth-fused reference point cloud for TUM.

On TUM, the standalone formulation obtains Chamfer-L1 error of $0.020$, close to \ac{GPDF} ($0.016$) and slightly better than the joint variant ($0.023$). On Cow\&Lady, it achieves $0.042$, improving over the joint variant ($0.055$) and outperforming \ac{GPDF} and \ac{2DGS}. Although HotSpot obtains low Chamfer-L1 errors, qualitative comparison shows that HotSpot produces oversmooth meshes. \ac{2DGS} fails to recover Cow\&Lady surfaces reliably, CAP-UDF fails to capture key details, particularly near the lady, and \ac{GPDF} inflates the objects, losing fine details. In contrast, our methods recover richer geometric detail across both datasets, albeit with some double surfaces near the lady.

 \subsection{Downstream Robotic Application: Navigation}
 
 \begin{figure}[!ht]
    \centering
    \setlength{\tabcolsep}{2pt}
    \setlength{\fboxsep}{0pt}
    \setlength{\fboxrule}{0.8pt}

    \begin{tabular}{@{}cc@{}}
        \fbox{\includegraphics[width=0.39\columnwidth,height=0.22\columnwidth]{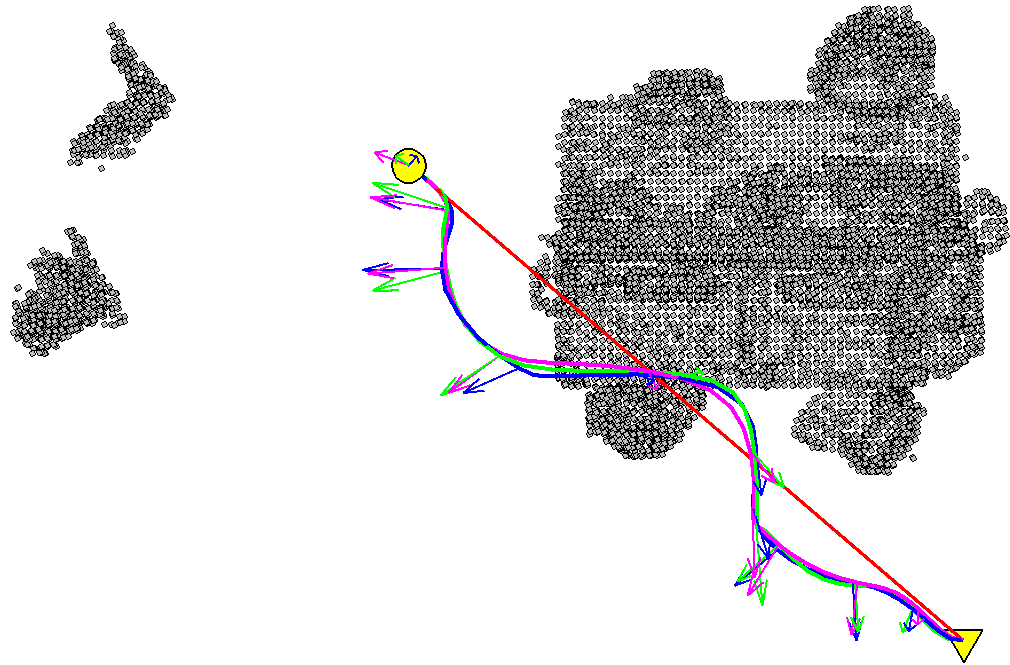}} &
        \fbox{\includegraphics[width=0.58\columnwidth,height=0.22\columnwidth]{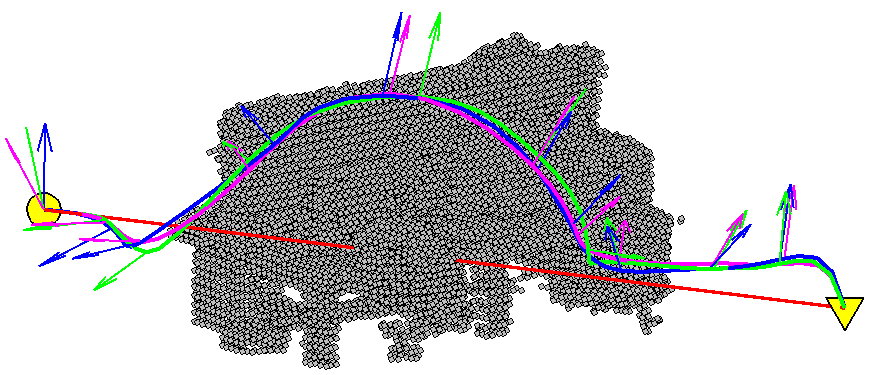}} \\[2pt]

        \fbox{\includegraphics[width=0.39\columnwidth,height=0.28\columnwidth]{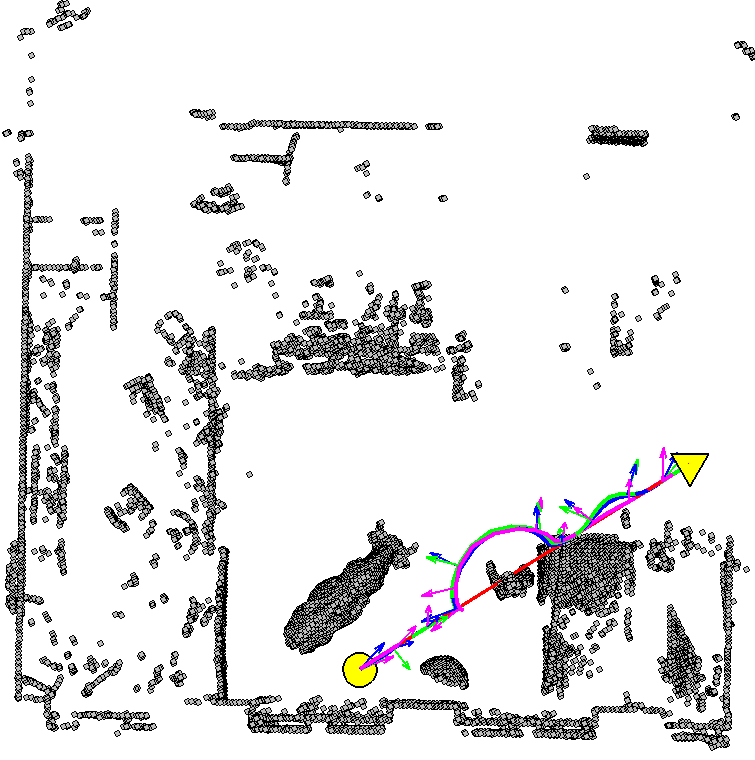}} &
        \fbox{\includegraphics[width=0.58\columnwidth,height=0.28\columnwidth]{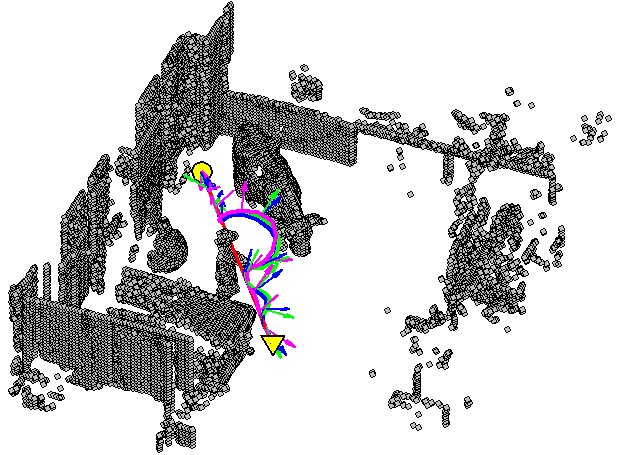}} \\[2pt]

        \scriptsize Top view & \scriptsize Side view \\[2pt]

        \multicolumn{2}{@{}c@{}}{%
    \fbox{%
        \includegraphics[width=0.94\columnwidth]{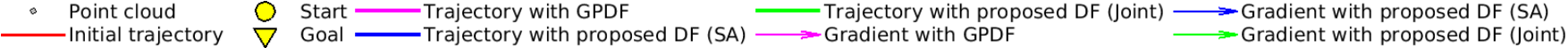}%
    }%
} \\
    \end{tabular}

    \caption{
    CHOMP trajectory optimisation on the TUM dataset (top row) and Cow\&Lady dataset (bottom row).
    Gradient arrows indicate directions away from the nearest surface. Ground and ceiling points have been removed for clarity.
    }
    \label{fig:tum_chomp_trajectory}
    \vspace{-1.5em}

\end{figure}

We evaluate the learned \acp{DF} in downstream navigation by using them as collision-cost fields for \ac{CHOMP}~\cite{CHOMP} on the 3D scenes. We assume a robot with a radius set to $r=0.20\,\mathrm{m}$ and a conservative clearance requirement of $0.25\,\mathrm{m}$. This clearance ensures that \ac{CHOMP} penalises waypoints such that this safety buffer is maintained from any obstacle along the full path. Since CHOMP leverages distance and gradient information to optimise the trajectory waypoints, accurate distance and gradient predictions are vital for navigation. We compare against \ac{GPDF}, the strongest-performing \ac{DF} benchmark in 3D after ours. The results in Figure~\ref{fig:tum_chomp_trajectory} show that all three methods yield similar guidance and smooth optimised trajectories with improved clearance. The predicted gradients push the initial colliding trajectory away from obstacles, ensuring safety-aware navigation on both scenes. This indicates that our proposed \acp{DF} are suitable for gradient-based path optimisation.

\section{Discussion and Conclusion}
\label{sec:discussion}

The \ac{DF} results support anisotropic Gaussian primitives as an effective parameterisation for continuous \ac{DF} mapping. In 2D, the standalone model remains competitive on snowflake and shows clearer gains on the larger Gazebo, indicating the benefit of tree-guided Gaussian allocation over broader free-space regions. In 3D, both variants achieve higher distance and gradient accuracy than the compared methods. The runtime comparison supports this trend, that the proposed frameworks scale favourably on larger datasets. Furthermore, although the proposed standalone method achieves the lowest RMSE, the joint formulation improves gradient alignment, with added capability of coupled rendering.
 
The rendering results indicate that coupling the \ac{DF} with \ac{2DGS} improves over vanilla \ac{2DGS} with higher PSNR and lower reconstruction losses. Qualitatively, the improvement is more visible on Cow\&Lady, where the joint formulation better preserves several scene details that vanilla \ac{2DGS} fails to capture.

The surface reconstruction results show that our methods, while not always best numerically, preserve important scene structures such as furniture legs in TUM and the key objects in Cow\&Lady. Some methods obtain lower Chamfer distances through smoother but less detailed reconstructions. The poor result in Cow\&Lady from \ac{2DGS} highlights the sensitivity of TSDF fusion to rendered-depth quality and image coverage. In contrast, by maintaining an explicit continuous \ac{DF}, our frameworks can do geometry-driven reconstruction, with the option to use \ac{2DGS}-style mesh reconstruction given reliable rendering. The navigation experiments further show that both formulations provide distances and gradients coherent enough for \ac{CHOMP}-based trajectory optimisation, enabling collision avoidance and safer navigation.
 
A limitation of the current formulation is that it predicts unsigned \acp{DF}, requiring an additional sign-recovery step for surface extraction with marching cubes. This makes reconstruction quality dependent on the reliability of the recovered sign field, especially in regions with sparse observations, noisy depth, or limited camera-ray coverage. Future work will investigate more robust sign recovery, stronger coupling, and online extensions for robotic mapping.

Overall, this work demonstrates that Gaussian primitives can provide continuous, differentiable and accurate \ac{DF}s capable of downstream surface reconstruction and navigation tasks. Additionally, unlike \ac{GPDF} methods, this formulation facilitates smoother integration with rendering frameworks, suggesting a promising direction for Gaussian-based scene representations that expose both visual appearance and robotics-oriented geometric queries.

\bibliographystyle{plain}
\bibliography{templates/reference}
\end{document}